\documentclass[12pt]{article}
\usepackage[a4paper, left=3.17cm, right=3.17cm, top=2.54cm, bottom=2.54cm]{geometry}
\usepackage[T1]{fontenc}
\usepackage{mathptmx}
\usepackage{amsmath}
\usepackage{amssymb}
\usepackage{amsfonts}
\usepackage{chemformula}
\usepackage{cite}
\usepackage[colorlinks, linkcolor=black, anchorcolor=black, citecolor=black]{hyperref}
\usepackage{graphicx}
\usepackage{float}
\usepackage{colortbl}
\usepackage{xcolor}
\usepackage{makecell}
\usepackage{multirow}
\usepackage{latex_template,times}
\title{Place Your Project Title at Here}
\title{%
  Google is all you need: Semi-Supervised Transfer Learning Strategy For Light Multimodal Multi-Task Classification Model
}

\nipsfinalcopy 
\begin{document}

\author{%
  Haixu Liu \\
  The University of Sydney\\
  \texttt{hliu2490@uni.sydney.edu.au} \\
  \And
  Zerui Tao \\
  The University of Sydney\\
  \texttt{ztao0063@uni.sydney.edu.au} \\
  \AND
  Penghao Jiang \\
  The University of Sydney \\
  \texttt{pjia0498@uni.sydney.edu.au} \\
}

\maketitle

\begin{abstract}
As the volume of digital image data increases, the effectiveness of image classification intensifies. This study introduces a robust multi-label classification system designed to assign multiple labels to a single image, addressing the complexity of images that may be associated with multiple categories (ranging from 1 to 19, excluding 12). We propose a multi-modal classifier that merges advanced image recognition algorithms with Natural Language Processing (NLP) models, incorporating a fusion module to integrate these distinct modalities. The purpose of integrating textual data is to enhance the accuracy of label prediction by providing contextual understanding that visual analysis alone cannot fully capture.

\noindent Our proposed classification model combines Convolutional Neural Networks (CNN) for image processing with NLP techniques for analyzing textual description (i.e., captions). This approach includes rigorous training and validation phases, with each model component verified and analyzed through ablation experiments. Preliminary results demonstrate the classifier's accuracy and efficiency, highlighting its potential as an automatic image-labeling system. The link of our code is \href{https://colab.research.google.com/drive/1IeTkVXBFNo8d78Gj0sx9PuFs0dTfOvcU?usp=sharing}{code link}.
\end{abstract}
\section{Introduction}
\subsection{Aim of Study}
This research aims to propose an effective and efficient multi-label image classification model that can assign one or more labels to each image by utilizing multi-modal features, ensuring high accuracy and feasible run time. More specifically, this work seeks to advance the field of image classification by investigating the integration of visual and textual information within a multi-modal framework. This approach is designed to harness the complementary strengths of visual and textual analysis, overcoming the limitations presented by employing either modality in isolation.

\noindent To develop a classifier with high accuracy and efficiency, this study focuses on exploring various feature extraction algorithms from the perspectives of image and text separately. It then investigates how to effectively fuse the features extracted from different modalities for the downstream classification task. Once the overall framework is established, the hyper-parameter tuning will be conducted to optimize the model's performance, followed by an ablation study to evaluate the contribution of each component to the model.


\noindent In sum, the efficacy of this multi-modal approach will be substantiated through rigorous training and validation phases. The classifier developed through this research could establish a benchmark for future explorations and applications in automated image processing and other related disciplines requiring advanced data analysis capabilities.

\subsection{Importance of Study}

Accurate automated image classification reduces the reliance on manually labeled images, saving significant time and resources. For instance, medical images can be categorized using image classification models to aid medical diagnoses. More specifically, multi-label classification task is addressing a more practical issue in real world, as a single instance is more often associated with multiple labels. For example, a photo can contain multiple objects, assigning it multiple tags (e.g., 'tree', 'car', 'bridge'). Such that, multi-label classification enables more accurate and comprehensive tagging by learning label correlations. Furthermore, when the model is capable of recognising the relationship among labels, it demonstrates its ability to capture more information and context about the data, resulting in improved performance. The enhanced understanding can help the model be more robust and reliable in various application.

\noindent Features learned from different modalities will complement each other in terms of the information they provide, allowing for a more comprehensive data representation when combined. The enriched multi-modal features will be more robust to noise and incomplete data, as compensation from other modalities can occur if one modality information is missing, enabling more reliable outputs.

\noindent To finalize the final best model, it is crucial to identify the most effective image and textual feature extractors by comparing various state-of-the-art (SOTA) algorithms before the multi-modal feature fusion step. Once the overall model structure with the feature fusion module is established, hyper-parameter tuning is necessary to figure out key parameters of the final model for optimal performance. In the end, it is also important to perform an ablation study accessing the necessity and impact of each component within the framework on the classification performance.


\subsection{General Introduction and Motivation}
We propose a multi-modal approach to construct an accurate and efficient multi-label image classification model. The method integrates three principal modules: the vision module, the natural language processing (NLP) module, and the feature fusion module.

\noindent \textbf{Vision Module: }This module extracts salient features from image data. It employs advanced image processing techniques and other methods based on convolutional neural networks (CNNs) to analyze and identify key visual patterns and structures within the images.

\noindent \textbf{Natural Language Processing Module:} The role of this module is to process text data, mining the structural and semantic nuances of the language to extract textual features. This module uses various NLP techniques such as tokenization, semantic analysis, and embedding generation to understand and quantify the textual context that complements the visual data.

\noindent \textbf{Feature Fusion Module:} The core function of this module is to integrate the outputs from both the vision and NLP modules. By fusing the extracted visual and textual features, this module enhances the model's capacity to assimilate information from diverse sources, thereby improving the accuracy and reliability of decision-making. This integration augments the classification process's precision and significantly boosts the model's generalization ability across unseen samples.

\noindent \textbf{Motivation:} Given image and textual information, we hope that our model can leverage the complementary strengths of both visual and textual information, contributing to a more robust and versatile classification system. The approach utilizing multi-modal data is pivotal in tackling the inherent complexities of multi-label image classification and sets a foundation for further exploration and enhancement in multi-modal machine learning applications.

\section{Related Works}

\subsection{Multi-Label Learning}
Deep learning-based methods designed for multi-label image classification tasks can be grouped into two: feature representation learning and label dependency learning. The former one aims to extract appropriate features from input data and then effectively connect them to related labels, thereby enhancing model performance. Recent successful feature-learning methods for the task are mostly based on convolutional neural networks (CNN)[15-17]. The latter, on the other hand, focuses on learning label relations by treating them as a sequence[18-20] or a graph[21-23]. 

\noindent With the availability of cross-modal data, multi-modal learning techniques have recently been integrated into either category to fuse the semantic information learned from texts (e.g., captions) with the spatial information extracted from images. Pre-trained language models are suggested as valuable semantic resources for multi-label image classification when combined with appropriate multi-modal learning techniques.

\noindent In this study, given inputs of images and captions, we focus on feature representation learning by integrating multi-modal learning techniques. Therefore, section 3.2 and 3.3 will demonstrate related works on visual and textual feature learning respectively.

\subsection{Visual Feature Learning}
With the development of deep learning, CNN architectures have become the basic of computer vision, introducing advanced models like AlexNet[1] and Inception[2] capturing essential spatial features for classifying images. These networks use multiple transformations, including convolutions and pooling, to detect and synthesize low-level and high-level features, although increasing complexity often comes at the cost of efficiency. Lightweight CNN architectures aim to streamline these models by reducing parameters and accelerating learning, maintaining accuracy while ensuring fast processing and low memory use, making them suitable for real-world applications with limited computational resources.

\noindent SqueezeNet[3], one of the first lightweight CNNs, matched the accuracy of AlexNet on ImageNet with 50 times fewer parameters, illustrating its suitability for resource-constrained environments. MobileNet[4], introduced by Google in 2016, is specifically designed for mobile devices and employs depth-separable convolutions, which reduce computational cost without compromising model effectiveness. This approach was further evolved in MobileNetV2[5] and MobileNetV3[6], which incorporated innovations such as the Inverted Residuals module from ResNet and the SENet (Squeeze and Excitation) mechanism, enhancing the model's attention capabilities.

\noindent ShuffleNet, developed by the Face++ team, includes iterations such as ShuffleNetV1[7] and ShuffleNetV2[8]. The first version introduces pointwise group convolution and channel shuffling, optimizing parameter efficiency and computational speed. The subsequent version, ShuffleNetV2, further refines these techniques, redesigning the basic unit architecture to optimize performance.

\noindent EfficientNet[9], another innovation by Google Research introduced in 2019, is based on a systematic study of model scaling. This model uses a novel composite scaling method to coordinate depth, width, and resolution scaling with fixed coefficients, enhancing model efficiency. The latest iteration, EfficientNetV2[10], introduced in 2021, offers improvements that make the model smaller and faster to train than its predecessors.

\subsubsection{MoblieNet}
MobileNet is a trendy lightweight neural network architecture developed by a Google team in 2017. It is designed to operate on resource-constrained devices such as mobile phones and other embedded systems. While its accuracy slightly trails that of VGG16—by approximately 0.9\%—its model size is a mere 1/32 of VGG16's. This impressive efficiency is primarily due to its use of Depthwise Separable Convolution (DW Convolution).

\noindent Traditional convolution operations involve each filter engaging with all channels of the input feature matrix, a process that requires significant computational resources. In contrast, depthwise separable convolution drastically reduces computational demands by splitting the traditional convolution into two phases: depthwise and pointwise convolution. In depthwise convolution, filters are applied separately to each input channel, which means each filter's channel dimension is set to 1, ensuring that the output feature matrix retains the same number of channels as the input.
To expand the number of output channels and boost the network's learning capacity, depthwise convolution is followed by pointwise convolution (PW Convolution) that employs 1x1 kernels. This sequence, known as depthwise separable convolution, substantially lowers the computational cost and the model size while preserving reasonable performance levels.

\noindent Moreover, this architecture dramatically reduces the number of model parameters by using separate convolution kernels for each input channel instead of having each convolution kernel span all input channels. However, one downside to depthwise separable convolution is that it diminishes the interaction between input channels, potentially impacting the network's learning capability. Nonetheless, this limitation can be effectively countered by fine-tuning the model architecture and optimizing training methodologies.

\noindent In this paper section, we will compare and elucidate the advantages of Depthwise Separable Convolution by contrasting it with Standard Convolution through a detailed mathematical analysis. Assume the input shape is \([D_{f}, D_{f}, M]\), the output shape is \([D_{f}, D_{f}, N]\), the kennel size is \([D_{k}, D_{k}]\), the computational cost of standard convolution would be detailed by Formula 1.
\begin{equation}
    D_k \cdot D_k \cdot M \cdot N \cdot D_f \cdot D_f
\end{equation}
In Formula 1, $D_k^2$ represents the size of the kernel, and M is the number of input channels. N is the number of output channels. $D_j^2$ corresponds to the total number of positions the kernel slides over (the spatial area of the output feature map).\\
Formula 2 and Formula 3 will detail the computational cost for depthwise separable convolutions. Formula 2 describes the process of deep convolution, where each input channel is individually convolved using a separate filter. Formula 3 describes the process of pointwise convolution. Pointwise convolution uses a 1x1 kernel to combine the outputs of the depth steps between channels, thus effectively amplifying the depth to match the desired N output channels.
\begin{equation}
    D_f^2 \cdot M \cdot D_k^2 
\end{equation}
\begin{equation}
    D_f^2 \cdot M \cdot N
\end{equation}
MobileNetV2 represents an iterative advancement from its predecessor, MobileNetV1, building upon its strengths while incorporating several key improvements to enhance performance and efficiency. This section details the central enhancements and modules integral to MobileNetV2.\\
\subsubsection{MobileNetV2}
\textbf{Inverted Residuals and Linear Bottlenecks:}
\noindent MobileNetV2 introduces a novel architectural element inverted residual blocks, which employ lightweight depthwise separable convolutions to filter features. A significant improvement in this architecture is the implementation of linear bottlenecks within the residual blocks. Unlike MobileNetV1, which utilized traditional residuals, MobileNetV2 incorporates linear activations in the terminal layer of each block, thereby preserving the representational power by maintaining the integrity of nonlinear activations. Furthermore, the inverted residual module is organized into three distinct stages: the extension, deep convolutional, and projection layers. 

\noindent \textbf{Extension Layer:} This layer increases the number of channels in the input feature map, creating a high-dimensional space conducive to processing the input data. The expansion factor, denoted by t, determines the extent of this increase. If the input possesses M channels, the expanded dimensionality becomes tM. Formulas 4 and 5 mathematically articulate the process of this layer. Assume $x \in \mathbb{R}^{H \times  W \times C}$ is the input tensor. $F_{expand}$ uses a set of filters $W_{expand} \in \mathbb{R}^{1 \times  1 \times C \times (ts)}$ and each filter's dimension is $1 \times 1$ to mapping C input channels into tC output channels. And h and w index the spatial dimensions. c indexes the input channels, k indexes the output channels, $b_{expand}$ is the bias term.
\begin{equation}
    x_{expanded}[h,w,k] = ReLU(\sum_{c =1}^{C}x[h,w,c] \cdot W_{expand}[1,1,c,k] + b_{expand}[k])
\end{equation}\
\begin{equation}
    Expand_Dimension = t \times M
\end{equation}
\textbf{Deep Convolution Layer: }This layer employs a 3x3 depthwise separable convolution on the expanded tensor, facilitating independent processing of the spatial elements within each channel. Such lightweight filtering preserves spatial resolution while minimizing computational costs. The operational specifics are delineated in Formula 6. We set depthwise convolutional filters $W_{dw} \in \mathbb{R}^{3 \times  3 \times tC \times 1}$ and applying to each channel. And i and j are offsets within the kernel window.
\begin{equation}
    x_{depththwise][h,w,k]} = \sum_{i=-1}^1 \sum_{j=-1}^1 x_{expanded[h + i,w+j ,k]} \cdot W_{dw}[i + ,j +1,k,1]
\end{equation}
\textbf{Projection Layer: }This layer projects the tensor post-deep convolution processing into a lower-dimensional space, utilizing linear activation to retain information from prior nonlinear transformations. The output from this layer aligns with the number of output channels N, which is typically smaller than the expanded dimension tM. The mathematical framework of this layer is outlined in Formula 7. $W_{project} \in \mathbb{R}^{1 \times  1 \times tC \times C}$ is the projection filter in the projection layer and c indexes the reduced output channels.
\begin{equation}
    x_{projected}[h,w,c] = \sum_{k=1}^{tC} x_{depthwise}[h,w,k] \cdot W_{project}[1,1,k,c] + b_project[c]
\end{equation}
Through these enhancements, MobileNetV2 significantly boosts efficiency and performance, making it a robust framework for mobile and embedded vision applications.

\noindent \textbf{Residual Connection:}

\noindent Residual concatenation enhances neural network architectures by allowing the output of one layer to be added directly to the input of a subsequent layer, thus creating shortcut paths that facilitate gradient flow during backpropagation. This technique is particularly effective in addressing the vanishing gradient problem, a pervasive challenge in deep learning networks where gradients propagated back through numerous layers diminish in magnitude, often resulting in slower convergence or suboptimal training outcomes.

\noindent The primary advantages of incorporating residual connectivity in neural networks are as follows:

\noindent \textbf{Improved Gradient Path:} Residual connectivity alleviates the vanishing gradient problem by providing a direct route for gradient accumulation during backpropagation. This enhancement ensures that gradients maintain their magnitude across layers, fostering more efficient learning processes.

\noindent \textbf{Ease of Training:} Networks equipped with residual connections generally train more swiftly and reliably than those lacking such features. This improvement is attributed to the sustained gradient magnitudes throughout the training phases, preventing the gradients from precipitously declining.

\noindent \textbf{Versatility and Flexibility: } Implementing residual connections enables the design of deeper neural network architectures capable of learning more intricate patterns. This adaptability is achieved without significantly elevating the risk of gradient vanishing, thus broadening the practical applications of such models in complex tasks.\\
The mathematical process of residual concatenation is detailed in Formulas 8 and 9. In these two forms, x represents the input to the block, and F(x) signifies the transformations executed within the block, including expansion, depthwise convolution, and projection. $F_{expand}(x)$ represents the number of expand channels. 
\begin{equation}
    x_{out} = x + F(x)
\end{equation}
\begin{equation}
    F(x) = F_{project}(F_{depthwise}(F_{expanded}(x)))
\end{equation}
\subsubsection{MobileNetV3}
MobileNetV3 inherits the features of MobileNetV2 and makes significant architectural and efficiency improvements. mobileNetV3 further improves the performance and efficiency of the model by incorporating the latest Network Architecture Search (NAS) technology and hardware-based optimizations.

\noindent \textbf{Network Architecture Search (NAS):}\\
Network Architecture Search (NAS) is a methodology that employs machine learning techniques to design and optimize neural network structures autonomously. In the development of MobileNetV3, NAS was utilized to refine the model's convolutional layers, enhancing efficiency and accuracy within predetermined hardware constraints. NAS optimizes these layers through a three-pronged approach: defining the search space, implementing a search strategy, and devising a performance evaluation strategy. The search space encompasses all conceivable network architectures, which, for MobileNetV3, include various types of convolutional layers, activation functions, convolutional kernel sizes, and shapes. The search strategy dictates the methodology for exploring this space. In contrast, the performance evaluation strategy assesses the effectiveness of a specific architecture, typically by executing a scaled-down version of the model on a validation set to gauge its performance rapidly. In MobileNetV3, NAS scrutinizes different convolutional kernel sizes and shapes to identify the optimal configuration that balances high accuracy with reduced computational complexity. Additionally, NAS automates decisions regarding the number of convolutional kernels per layer, connectivity options—including residual connections or Squeeze-and-Excitation (SE) modules—and other architectural choices. The objective function $\mathbb{J}(\theta,\alpha)$ drives the optimization process. $\theta$ is the model parameters and $\alpha$ is the hyperparameter which is described the network structure.$L(\theta,\alpha)$ is the loss function of model parameter $\theta$ for a given architecture $\alpha$. The math process is detailed mathematically in Formula 10.
\begin{equation}
\min_{\alpha} \quad J(\theta^*(\alpha), \alpha) \quad \text{where} \quad \theta^*(\alpha) = \arg\min_{\theta} L(\theta, \alpha)
\end{equation}
\textbf{H-Swish Activiation Function:}\\
H-Swish is an activation function employed in deep learning models, serving as a hybrid variant of the Swish and ReLU activation functions. Designed to be hardware-friendly, H-Swish aims to deliver performance benefits comparable to Swish while maintaining computational efficiency. The primary function of H-Swish is to introduce nonlinearities within neural networks, particularly to ensure optimal performance on resource-constrained devices. The mathematical formulation of the H-Swish activation function is specified in Formulas 11 and 12.
\begin{equation}
    h(x) = x \cdot \frac{ReLU6(x + 3)}{6} 
\end{equation}
\begin{equation}
    ReLu6(x) = min(max(x,0),6)
\end{equation}
H-Swish facilitates smooth activation, exhibiting near-linearity in positive intervals. This characteristic aids in maintaining network stability during training while simultaneously enabling the learning of deeper features. Such properties are crucial for practical training of complex neural network architectures, as they ensure a steady learning progression without the instability typically associated with non-linear activation functions.

\noindent \textbf{Squeeze-and-Excitation Block:}\\
The Squeeze-and-Excitation (SE) module represents a significant enhancement in deep learning architectures aimed at boosting network performance by recalibrating the feature channels of the convolutional layer. This recalibration enhances the model's focus on salient features, improving overall accuracy. The SE module operates through two principal steps: Squeeze and Excitation.

\noindent The squeeze step aggregates global information from each feature channel. Conventionally, this aggregation is performed using global average pooling after the convolutional layer. Precisely, for each feature channel, the average across all spatial dimensions (i.e., width and height) is calculated to derive a single scalar value per channel. This scalar effectively summarizes the global distribution of the channel across the entire input feature map. The mathematical formulation for this compression process is provided in Formula 13. $z_c$ represents the global average value of channel c. H is the feature map's height, and W is the feature map's width.$x_c{i,j}$ represents the eigenvalue of location (i,j).
\begin{equation}
    z_c = \frac{1}{H \times W} \sum_{i=1}^H \sum_{j=1}^W x_c(i, j)
\end{equation}
The excitation step models the significance of each channel using learned weights. The core of this step is a compact, fully connected neural network. The process begins with the network passing through a dimensionality-reduction fully connected layer, reducing the number of parameters and curtailing computational complexity. A ReLU activation function follows the process. Subsequently, the network passes through another fully connected layer where the dimensionality is increased, and a sigmoid activation function is applied to generate weight coefficients for each channel. These coefficients reflect the channel's importance relative to the specific task. The mathematical expression for the excitation step is detailed in Formula 14. z is the extruded channel descriptor. $W_1$ and $W_2$ are the weights of the fully connected layer. $s_c$ is the incentive weight of channel c.
\begin{equation}
    s_c = Sigmoid(g(z,W) = Sigmoid(W_2\cdot ReLU(W_1 z))
\end{equation}
\subsubsection{ShuffluNet}
ShuffleNet is a highly efficient Convolutional Neural Network (CNN) architecture designed for devices constrained by computational resources. A key attribute of ShuffleNet is its capacity to deliver performance on par with more complex networks while operating at significantly lower computational complexity. ShuffleNet achieves reductions in computational complexity and parameter count, maintaining robust network performance through two primary mechanisms: grouped convolution and channel shuffling.

\noindent \textbf{Grouped Convolution:}\\
AlexNet initially proposed grouped Convolution to address GPU memory constraints, and grouped Convolution has been extensively utilized in ShuffleNet to decrease the model's computational burden and parameter count. Unlike traditional convolutional operations, where each element of the input feature map interacts with all convolutional kernels, grouped Convolution mitigates this computationally intensive requirement by dividing the input feature map and convolution kernels into multiple independent groups. Each group of kernels conducts convolution operations solely with a corresponding segment of the input feature map, thereby significantly reducing the number of multiplicative operations.

\noindent Assume the size of the input feature map \(X\) is \((H, W, C_{\text{in}})\), where \(H\) and \(W\) are the height and width of the feature map are feature maps respectively, \(C_{\text{in}}\) is the number of input channels. If we use \( g \) groups for convolution, each group processes \(\frac{C_{\text{in}}}{g}\) input channels. If we want the number of output channels per group to be \( C_{\text{out}} \), then the number of output channels per group should be \(\frac{C_{\text{out}}} {G} \). The mathematical formulation for grouped convolution is detailed in Formula 15. $Y_k^{(g)}$ is the kth output feature map of the g-th group. $X^{(g)}$ is the part of the input feature map in group g. $W_k^{(g)}$ is the convolution kernel of the g-th group for the kth output feature map, and * is the convolution operation.
\begin{equation}
    Y_k^{(g)} = X^{(g)} * W_k^{(g)}
\end{equation}
\begin{figure}[H]
        \centering
        \includegraphics[width=\textwidth]{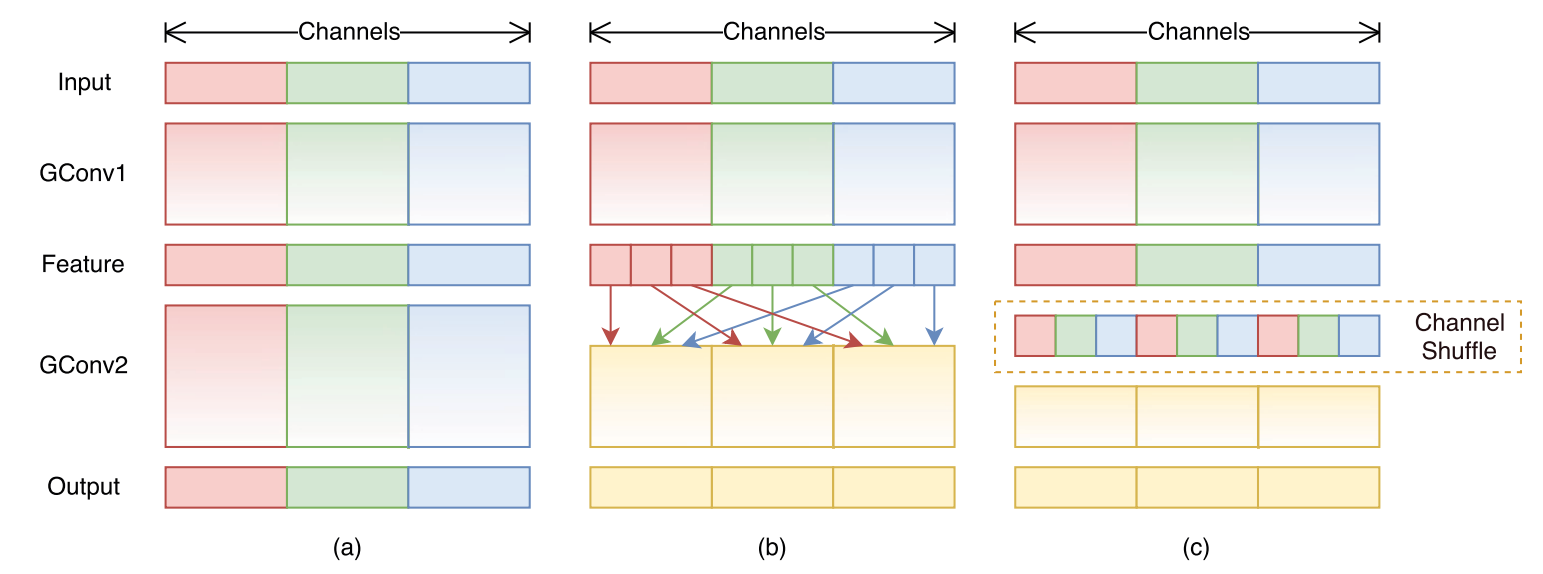}
        \caption{Channel Shuffle}
\end{figure}
\noindent \textbf{Channel Shuffle:}
Channel Shuffle addresses this issue by recombining the channels from the output feature maps of the group convolutions. After performing the group convolution, the Channel Shuffle operation rearranges the output channels across the groups. This rearrangement ensures that the subsequent layers receive a mix of features from all the previous groups, thus restoring inter-group communication and helping to preserve and distribute information more effectively throughout the network. After shuffling, another set of pointwise group convolutions is applied, which processes this richer, mixed set of features. By enabling this cross-group information flow, ShuffleNet effectively combats the limitations of isolated group convolutions, making it a powerful yet efficient architecture for deep learning tasks in constrained environments.

\noindent Assume an output layer has \( g \) groups, each with \( n \) channels, making the total number of channels \( C = g \times n \). The output feature map can be denoted as \( X \in \mathbb{R}^{H \times W \times C} \). Channel shuffling aims to rearrange the order of the \( C \) channels.
Suppose \( X \) is divided into \( g \) groups, each containing \( n \) channels, i.e., \( X = [X_1, X_2, \dots, X_g] \), where \( X_i \) represents the output feature map of the \( i \)-th group. After channel shuffling, we obtain a new feature map \( Y = [Y_1, Y_2, \dots, Y_g] \), where each \( Y_i \) contains channels from all groups in \( X \).
A common implementation method is to shuffle each group's channels based on the index \( i \) using modulo \( g \) operations and then recombine them. In other words, the \( j \)-th channel of the \( i \)-th group, \( Y_{ij} \), is derived from \( X_{(i+j) \mod g, j} \).

\noindent ShuffleNetV2 advances beyond simply optimizing FLOPs by addressing additional aspects that influence network execution speed, such as the drawbacks of excessive group convolutions. Expanding on the foundational design of the ShuffleNetV1 block, ShuffleNetV2 integrates a "Channel Split" mechanism at the beginning of each block. This process divides the input feature map into two distinct branches: one is preserved without alteration. At the same time, the other undergoes processing through a revised ShuffleNetV1 block that omits channel shuffling, given that the initial split has already segregated the channels effectively.

\noindent Subsequently, the outputs from both branches are merged via concatenation. This design facilitates a pattern of feature reuse akin to that observed in DenseNet, leading to enhanced accuracy and efficiency. The architectural modifications in ShuffleNetV2 reduce computational complexity and improve the overall performance of the network, making it more adept at handling resource-intensive tasks in environments with limited computational capabilities.
\subsubsection{EfficientNet}
EfficientNet is an efficient convolutional neural network architecture proposed by Google in 2019, designed to optimize the model's performance and efficiency through compound scaling. Its primary feature is the simultaneous adjustment of the input image's depth, width, and resolution to achieve optimal performance under various computational resource constraints. The core idea of EfficientNet is to scale neural networks using compound scaling strategies systematically. Traditional approaches typically extend the network's depth, width, or resolution independently, which can lead to imbalanced models. EfficientNet's compound scaling method adjusts all three dimensions simultaneously to achieve a more balanced and efficient network expansion.

\noindent Increasing the depth in a deep learning network allows the model to capture more features, thereby enhancing its expressive ability. However, as the depth increases by d times, the computational complexity increases by d times. Increasing the width improves the feature representation ability of each layer, with the computational complexity increasing by w times when the width is scaled by $w^2$ times. Enhancing the resolution boosts the model's ability to capture detailed features, and when the resolution increases by r times, the computational complexity also rises by $r^2$ times.

\noindent The basic principle of the compound scaling method is to balance the three key dimensions of the network (depth, width, resolution) by adjusting a scaling factor $\phi$, rather than expanding one of the dimensions alone. This approach ensures that the network scales evenly across all dimensions, thereby improving model performance while maintaining computational efficiency. The mathematical expressions of the compound scaling method are detailed in formulas 16,17 and 18. $\alpha$, $\beta$, and $\gamma$ are fixed constants representing the basic scaling ratios of depth, width, and resolution, respectively. The scaling factor $\phi$ is used to control the expansion ratio of the model.
\begin{equation}
    depth = \alpha^{\phi}
\end{equation}
\begin{equation}
    width = \beta^{\phi}
\end{equation}
\begin{equation}
    resolution = \gamma^{\phi}
\end{equation}
\textbf{Goal Function:}
The goal is to find the optimal values for $\alpha$, $\beta$, and $\gamma$ such that, given a fixed computational budget (e.g., fixed FLOPs), the model accuracy is maximized. We first define an objective function to represent the computational complexity (FLOPs) with formula 19 and 20.
\begin{equation}
    \text{FLOPs} \propto d \cdot w^2 \cdot r^2
\end{equation}
\begin{equation}
    \text{FLOPs} \propto (\alpha^\phi) \cdot (\beta^\phi)^2 \cdot (\gamma^\phi)^2 = (\alpha \cdot \beta^2 \cdot \gamma^2)^\phi
\end{equation}
To ensure that the computational complexity does not exceed the budget, we need to make $\alpha$, $\beta$, and $\gamma$ meet the constraints as formula 21 detailed where $K$ is a constant representing the upper limit of the computational budget.
\begin{equation}
    \alpha \cdot \beta^2 \cdot \gamma^2 = K
\end{equation}

\subsubsection{EfficientnetV2}
EfficientNet represents a highly efficient Convolutional Neural Network (CNN) architecture that optimizes network performance through a systematic approach to scaling, which involves adjusting the depth, width, and resolution simultaneously. EfficientNetV2 builds upon the foundations of EfficientNet by incorporating three innovative techniques and strategies to augment efficiency and effectiveness further.

\noindent \textbf{Use of Lighter Weight Convolutional Operations:}\\
EfficientNetV2 introduces the Fused-MBConv convolution technique, simplifying the traditional MBConv block. The critical innovation of Fused-MBConv lies in its ability to 'fuse' specific convolution operations together, thereby reducing the number of layers and computational complexity. Typically, Fused-MBConv combines a 1x1 expansion convolution with a subsequent deep convolution into a single regular convolution operation, such as a 3x3 convolution. This integration eliminates the need for separate expansion layers and deep convolutions, streamlining the module, substantially lowering computational complexity, and enhancing computational efficiency.

\noindent \textbf{Using a Progressive Training Strategy:}\\
EfficientNetV2 employs a progressive training strategy, beginning the training process at a low resolution and incrementally increasing the resolution. This approach accelerates training speed and reduces memory consumption during training. The gradual transition from low to high resolution enables the model to enhance its learning from general to specific features. In the initial phases, the model captures more abstract and fundamental image features; as the resolution increases, it refines these features and adjusts to more intricate details, thereby improving the model’s generalization ability and overall performance.

\noindent \textbf{Improved Scaling Method:}\\
The scaling strategy of EfficientNet involves simultaneous adjustments to the depth, width, and input resolution of the network, guided by compounding coefficients denoted as 
$\phi$. This method facilitates a systematic enhancement of the network’s capacity and performance. The mathematical expressions for this initial scaling are detailed in Formula 22, 23, and 24. In these formulas, $d_0$ represents the number of layers in the basic network. $w_0$ represents the width of the network and $r_o$ represents the resolution.$\alpha$ represents the depth scaling factor, which controls the growth of network layers.$\beta$ represents the width scaling factor, which controls the growth of the number of channels in each network layer. $\gamma$ represents the resolution scaling factor, which controls the growth of the input image size, and $\phi$ represents the compound scaling coefficient, allowing users to adjust this value to scale the entire network as needed.
\begin{equation}
    d = d_0 \times a^{\phi}
\end{equation}
\begin{equation}
    w = w_0 \times \beta^{\phi}
\end{equation}
\begin{equation}
    r = r_0 \times \gamma^{\phi}
\end{equation}
EfficientNetV2 enhances the original scaling methodology to adapt more efficiently to diverse training conditions and application demands. It revises and optimizes the foundational scaling parameters of EfficientNet, achieving a more refined balance between computational efficiency and model performance. This optimization allows EfficientNetV2 to manage varying types and sizes of datasets efficiently.

\noindent Specifically, EfficientNetV2 excels in processing images of different resolutions and complexities. This capability stems from its advanced scaling strategy, which not only adjusts the depth, width, and resolution based on the application's specific needs but also optimizes these parameters to handle the diverse characteristics of datasets effectively. It makes EfficientNetV2 particularly adept at dealing with a wide range of image processing tasks, from those requiring high-resolution detail to those demanding rapid processing of less complex images, enhancing the model's versatility and applicability in real-world scenarios.
\subsection{Textual Feature Learning}
Word embedding offers a technique for representing words in a continuous vector space, encapsulating semantic and syntactic relationships. One of the seminal models, the Word2Vec[11], was developed by Google in 2013 and employs neural networks to discern word associations. This model features two primary components: Continuous Bag of Words (CBOW) and Skip-Gram. CBOW predicts a target word based on contextual words, whereas Skip-Gram uses a word to predict its surrounding context. These methods are particularly effective in capturing word relationships and enhancing the processing efficiency for large datasets.

\noindent Another notable approach, GloVe (Global Vectors for Word Representation)[12], developed by Stanford University, merges the benefits of matrix factorization methods (as seen in latent semantic analysis) with the contextual learning strategies employed by Word2Vec. GloVe constructs a co-occurrence matrix, recording the frequency of word pair occurrences within a specified context window. It then employs matrix decomposition to produce more condensed word vectors.

\noindent In a significant advancement in 2018, Google introduced Bert (Bidirectional et al. from Transformers)[13]. Utilizing the Transformer architecture, Bert has significantly impacted the Natural Language Processing (NLP) field. Differing from earlier models, Bert generates context-sensitive embeddings by considering both left and right context across all layers, which helps the model to grasp more complex linguistic structures effectively.

\noindent However, as these models grow in complexity and size, they face increasing challenges, such as GPU/TPU memory constraints and prolonged training periods. To mitigate these issues, Google developed ALBERT (A Lite BERT)[14] in 2020. ALBERT refines Bert's architecture by sharing parameters across different layers and reducing vocabulary size, substantially lowering memory demands and boosting training efficiency. Despite its streamlined design, ALBERT achieves performance comparable to its predecessor, Bert, making it more suitable for practical training and deployment scenarios.
\subsubsection{Bert(Tiny)}
BERT (Tiny) is a streamlined version of the BERT (Bidirectional et al. from Transformers) model, engineered to minimize computational resource demands while achieving relatively high performance. Unlike the standard BERT models, where BERT Base comprises 12 transformer layers, and BERT Large contains 24, BERT (Tiny) features only 2-4 layers of transformers. Moreover, each transformer layer in BERT (Tiny) includes fewer hidden units and self-attention heads, typically ranging from 2-4 heads, compared to the 12 heads found in BERT Base.

\noindent The underlying architecture of BERT (Tiny) is based on the transformer framework, which utilizes the self-attention mechanism to simultaneously process all elements of a sequence. This approach significantly enhances processing speed and efficiency. Central to the BERT model is its ability to comprehend and represent the contextual relationships within textual data.

\noindent \textbf{Self-attention mechanism:}
\noindent The core function of the self-attention mechanism is to enable the model to process a sequence while recognizing the interdependencies among the elements within that sequence. Each word in the input sequence is initially converted into a fixed-size vector representation. This vector can be sourced from pre-trained word embeddings or generated internally by the model. These vectors undergo various linear transformations to produce three distinct sets of vectors: query (Q), key (K), and value (V). The attention scores are computed using these vectors, where self-attention is achieved by calculating the dot product between each query and all keys, providing a measure of each element's influence over others in the sequence. The scores are typically scaled down following the dot product computation to mitigate the risk of gradient vanishing during training. The process for calculating attention is delineated in Formula 25. $\sqrt{d_k}$is a constant used to scale the result of the dot product to prevent gradient vanishing or explosion during training. The softmax function is applied to each row to convert the dot product score to a probability.
\begin{equation}
    \text{Attention}(Q, K, V) = \text{softmax}\left(\frac{QK^T}{\sqrt{d_k}}\right)V
\end{equation}
This streamlined design of BERT (Tiny) allows for efficient model training and deployment, especially in environments with strict computational constraints, without severely compromising the model's ability to handle complex language understanding tasks. BERT (Tiny) is a model optimized for environments with limited computational resources while maintaining the advantages of the core BERT technology. It provides a trade-off between computational cost and performance in practical applications.
\subsubsection{ALBERT}
ALBERT (A Lite BERT) is an optimized version of the BERT model, specifically designed to enhance memory efficiency and training speed by significantly reducing the number of model parameters. Like BERT, ALBERT is built upon the Transformer architecture and utilizes the self-attention mechanism for text processing. However, ALBERT introduces two major innovations to decrease parameter count and improve memory usage: the parameter sharing mechanism and factorized embedding parameterization.

\noindent \textbf{Parameter Sharing Mechanism:}

\noindent Unlike traditional BERT, where each layer maintains independent parameters necessitating separate updates and storage during training, ALBERT employs a parameter-sharing strategy. This approach involves reusing the same parameters across all Transformer layers, including those of the self-attention layers and the feed-forward network layers. Consequently, weights and biases learned in one layer are applied directly to subsequent layers. This mechanism ensures that the model's memory footprint is significantly reduced and acts as a form of regularization. By limiting the degrees of freedom, parameter sharing helps prevent overfitting, which is particularly beneficial when training data are limited..

\noindent \textbf{Factorized Embedding Parameterization:}

\noindent In conventional BERT models, the word embedding layer, which transforms words into dense vector representations to capture semantic relationships, typically involves many parameters due to large vocabularies. ALBERT addresses this issue with factorized embedding parameterization, where the traditional high-dimensional word embedding matrix is decomposed into two smaller matrices. This decomposition reduces the parameter count while maintaining model performance. In ALBERT, the embedding process involves two matrices: E and R. Matrix E maps words from the vocabulary index to a lower-dimensional embedding space, and matrix R maps these embeddings to the model’s hidden layer size. The resulting embedding for a word w can be mathematically represented as shown in Formula 26, where $E[w]$ is the embedding of the word $w$ in $E$, and $R$ is the transformation matrix applied on $E[w]$ to project it to the hidden layer dimension.
\begin{equation}
    h = R \times E[w]
\end{equation}
However, the parameter compression operation in ALBERT may lead to a slight performance trade-off compared to models without such compression. Additionally, using a parameter-sharing method in ALBERT necessitates more meticulous hyperparameter tuning to optimize model performance. This detailed tuning process can be time-consuming and labor-intensive, requiring careful exploration and validation of hyperparameter settings. As a result, while ALBERT provides significant efficiency and memory usage benefits, it may also demand more significant effort during the tuning phase to achieve the best possible outcomes.

\section{Technology}

The overview of our proposed method is shown in Figure 2. Detailed explanation and justification on each component of our proposed model will be explained and discussed in the following subsections (i.e., section 4.1 introduces the overall framework and section 4.2 introduces the novel fusion module).
\begin{figure}[H]
        \centering
        \includegraphics[width=\textwidth]{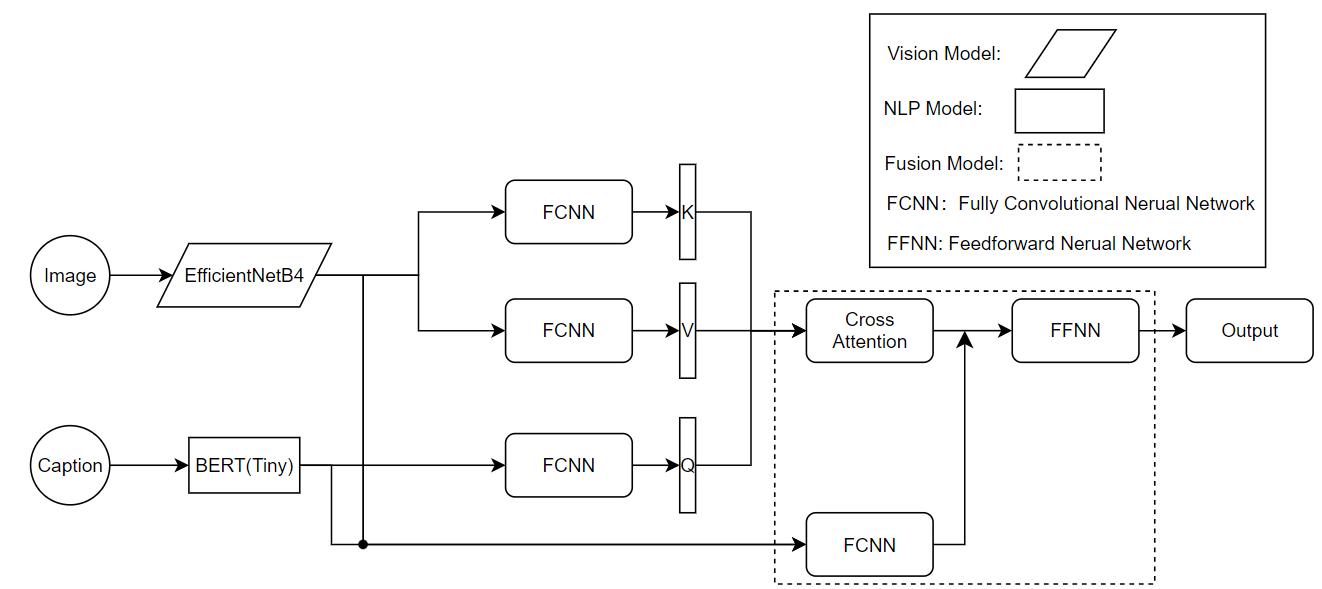}
        \caption{Overview of Our Proposed Method}
\end{figure}
\subsection{Interpretation of Our Method}
To fully leverage the dataset's rich information for this multiclassification task, we developed a multi-modal model composed of specialized modules for visual processing, natural language processing (NLP), and a fusion module that integrates outputs from the first two. Considering the constraints and requirements of the task, we selected three lightweight visual models for our visual processing module —EfficientNet, MobileNet, and ShuffleNet—due to their efficiency in environments with limited computational resources. For the natural language processing component, we opted for Tiny-BERT and ALBERT, which are similarly optimized for performance in resource-constrained settings. These models are characterized by their reduced parameter count and notably faster training speeds compared to traditional deep-learning models

\noindent For the Fusion Module of our multi-modal model, we employ a strategy of direct logit fusion to integrate outputs from the vision and natural language processing (NLP) modules. Principles of transfer learning underpin this process. Specifically, we freeze the weights of both the vision and NLP modules to utilize pre-trained efficiencies and minimize overfitting during the fusion process. The extracted feature vectors from each module are combined using a fully connected layer, which serves as the fusion point. The combined features are subsequently transformed into logits, which are the final outputs representing the integrated data from both modalities.

\noindent Additionally, we have designed a cross-attention module that incorporates a residual layer. This module inputs the combined feature vectors from the vision and NLP modules. The cross-attention mechanism focuses on enhancing feature integration by dynamically adjusting the attention to more relevant features from each module based on the task at hand. The output from this cross-attention module is also processed through fully connected layers to produce logits.

\noindent The inclusion of a residual layer within the cross-attention module provides significant advantages. It helps to mitigate the vanishing gradient problem by allowing gradients to flow through the network more effectively during backpropagation. This operation benefits training stability and can lead to more robust learning outcomes. Moreover, the residual layer supports deeper network architectures without the risk of performance degradation, thereby enhancing the model's overall performance. This fusion approach, combining direct logit fusion with advanced attention mechanisms and residual learning, aims to maximize the synergistic effects of visual and textual data, thus optimizing the model's performance on multiclassification tasks.

\subsubsection{Data Augumentation Method}
To enhance the performance of our multi-modal model, we implemented data augmentation techniques on both textual and visual components of the dataset. Our primary concern was that several images were associated with identical captions, potentially limiting the model's ability to learn distinctive features from the textual data. To address this, we enriched the text modality by augmenting the captions within the dataset.

\noindent We increased the variability of the captions by altering the vocabulary. Specifically, we randomly replaced two words in each caption with synonyms sourced from a predefined synonym list. This method diversifies the captions and enriches the textual information available for training, aiding the model in learning more general linguistic features. Such augmentation helps broaden the model's understanding and responsiveness to different textual expressions of similar concepts.

\noindent For the imaging modality, we employed horizontal flipping as our augmentation strategy. The model is encouraged to learn features invariant to horizontal orientation by randomly flipping images horizontally during training. This augmentation increases the robustness of the model against variations in image orientation and enhances the generalization capabilities of the network. Additionally, this technique has been observed to expedite the convergence of the network during training, making the learning process more efficient.

\noindent These data augmentation strategies collectively aim to create a more robust and versatile model by providing a richer and more varied training data set. By integrating these enhancements, we anticipate improvements in the model's accuracy and ability to generalize across different and unseen data scenarios.
\subsubsection{Weighted Loss Fuction}
To address the issue of label imbalance in multi-class classification tasks, we developed a novel weighting method based on logarithmic operations. We calculate the logarithm of the total number of labels, denoted as \( \log(T) \), and the logarithm of the number of labels for each category, denoted as \( \log(n_i) \). Two weight vectors are generated: the first vector assigns weights as \( \frac{\log(n_i)}{\log(T)} \), favoring categories with a more significant number of labels, while the second vector uses \( \frac{\log(T)}{\log(n_i)} \), which gives higher weights to categories with fewer labels. By averaging these two vectors, a new weight vector is formed where both minority and majority classes receive substantial weights, whereas categories with moderate numbers of labels receive lesser weights. This approach enhances the prediction performance of minority classes without significantly sacrificing the accuracy of the majority classes.
\subsubsection{Label Assignment Strategy}
During the model evaluation and testing phase, we encountered an issue where some model outputs displayed empty labels despite no test images explicitly assigned empty labels in the dataset. This issue stemmed from the model's classification mechanism, where each category's output probability was determined via a sigmoid function, with a standard threshold of 0.5 set for positive classification. Consequently, if no category probabilities exceeded this threshold, the sample would receive no labels, resulting in "empty labels."

\noindent To address this, we revised our label assignment method. For samples that would have previously resulted in empty labels, we now assign the label of the category with the highest probability, regardless of whether this probability surpasses the 0.5 threshold. This adjustment ensures that each sample is assigned a label, thereby eliminating instances of empty labels.

\noindent This new strategy not only resolves the issue of high classification thresholds leading to unassigned labels but also enhances the model's robustness and accuracy in practical applications. By ensuring that every sample is labeled, we can maintain consistency and reliability in the model's output, which is crucial for downstream tasks and real-world applicability. This approach also prevents the model from disregarding weaker signals that may still provide valuable predictive insights despite not meeting the stringent threshold.

\subsubsection{Special Train Strategy}
We find that a significant improvement in F1 scores is obtained by averaging the logits obtained for the output of the visual and natural language processing modalities. This situation suggests that the models of the two modalities label the wrong images differently, while the output obtained by averaging the logits contains information that is difficult to learn for the models of the two modalities, allowing the final model to take advantage of the information that is unique to each modality, thus improving the overall classification performance.

\noindent Inspired by these findings, we ventured into a semi-supervised learning approach to exploit this multi-modal synergy further. By treating the labels predicted by our model for the test dataset as pseudo-labels, we effectively increased the volume and diversity of training data. We merged this augmented dataset, comprising both the original and test data with pseudo-labels, to create a new training set.

\noindent We then trained two models on this enriched training set, one for each modality, allowing each to learn more comprehensive feature representations. After training, the logits from these newly trained models were fused to generate refined and more accurate pseudo-labels for the test set. This cycle of semi-supervised training was repeated over several epochs until there was no further improvement in the F1 score on a held-out validation set.

\noindent This iterative process underscores the utility of unlabeled data when appropriately leveraged and demonstrates the power of multi-modal fusion strategies in enhancing model generalization. The approach has shown that innovative training strategies and fusion of different modal outputs can substantially elevate the performance and robustness of classification models in complex scenarios. This advancement in training methodology provides a promising avenue for leveraging unexploited data to maximize the informational gains from each modality, thereby refining model accuracy and reliability in practical applications.

\subsection{Fusion Module}
In our fusion module design, we combined the logits from the visual and the natural language processing (NLP) modules, employing the EfficientNet\_b4 model for visual data and the BERT (Tiny) model for NLP tasks. The initial strategy involves freezing the weights of these two models during the training phase. Freezing weights is a common practice used to harness pre-trained models' robust feature extraction capabilities without altering their learned behaviors. It is particularly beneficial when training data is limited and the risk of overfitting is high.

\noindent Following the freezing of weights, the feature vectors extracted by the visual and NLP modules are concatenated. This method of splicing together the outputs from two distinct data types—visual and textual—aims to capitalize on the unique strengths of each modality. By integrating these diverse features, we enhance the model's ability to capture a broader spectrum of information, potentially leading to superior predictive performance in complex multi-classification tasks. Assuming the text input is T and the image input is I. The output feature of EfficientNet\_b4 is$F_I$ which could be detailed by Formula 27, and the output feature of BERT is $F_T$, which could be detailed by Formula 28.
\begin{equation}
    F_t = BERT(T)
\end{equation}
\begin{equation}
    F_I = EfficientNet_b4(I)
\end{equation}
After extracting the feature vectors from the EfficientNet\_b4 model for the visual module and the BERT (Tiny) model for the natural language processing module, we combine these vectors through a feature splicing operation. This operation is crucial as it merges the distinct modalities, leveraging their respective strengths to enhance the overall model performance. The mathematical expression for the splicing operation is shown in Equation 29, and the symbol $;$ represents the splicing operation.
\begin{equation}
    F = [F_T ; F_I]
\end{equation}
The combined feature vector fed into a fully connected (FC) layer. This layer serves as a dense layer in the neural network architecture, where each input node is connected to each output node and is primarily used for mapping the spliced features to the desired output space. The mathematical expression for the operation performed by this fully connected layer can be described as formula 30, where W is the weight of the fully connected layer and b is bias.
\begin{equation}
    L = W \cdot F + b
\end{equation}
Incorporating cross-attention mechanisms with residual layers into the feature fusion module is a strategic enhancement to refine the interplay between our multi-modal model's visual and natural language processing (NLP) modalities. This integration addresses the challenges of correlating distinct data types, which is crucial for tasks involving complex data interactions.\\

\subsubsection{Logits Fusion}
Logits fusion is an effective technique in multimodal learning, particularly when combining the strengths of models trained on different types of data or modalities. The logits, which are the outputs of models before applying a softmax or similar activation function, effectively represent raw, unnormalized scores that can be interpreted as the model's confidence in each class of a classification task. The model can leverage a richer information set to make more accurate predictions by fusing these logits from multiple sources.\\
Assuming our multimodal model operates with two different modalities, visual and textual, the output logits from the visual modality can be represented as $logits_1 = [z_{11},z_{12},...z_{1k}$ and from the textual modality as $logits_2 = [z_{21},z_{22},...z_{2k}$. We then fuse these logits using simple averaging. The mathematical expression for this fusion process is detailed in ForFormula 31.
\begin{equation}
    logits_{fusion} = \frac{1}{2}(logits_1 + logits_2)
\end{equation}
Formula 31 indicates that each element of the fused logits vector is the average of the corresponding elements from both the visual and textual output logits. This simple averaging method effectively combines the distinct strengths of each modality, potentially enhancing the model's overall predictive performance.\\
When the logits from the visual and textual modalities are fused using simple averaging, an activation function such as softmax is applied to convert these combined logits into a final predicted probability distribution. This conversion is crucial for interpreting the logits as probabilities, which indicate the likelihood of each category being the true class for a given input. The mathematical expression for applying activation function is detailed in Formula 31 where $z_j$ is the value of class j in the fused logits.
\begin{equation}
    P(y=j|\mathbf{x}) = \frac{e^{z_j}}{\sum_{i=1}^k e^{z_i}}
\end{equation}
\subsubsection{Cross-Attention Mechanism}
Cross-attention facilitates a dynamic focus on the most pertinent aspects of one data sequence about another. In the context of our model, cross-attention allows the system to allocate attention selectively across the visual and text features, emphasizing the elements most relevant to the task at hand. By computing the attention distribution from the visual to the textual sequence (or vice versa), the model can enhance its interpretative accuracy, focusing computational resources on analyzing and integrating the most informative features from both modalities.

\noindent We observed that many images with different labels correspond to the same caption. We also noticed that the encoding vector for text is 128 dimensions, while that for images is 1792 dimensions. From this, we deduce that the amount of information carried by text is smaller than that carried by images. Based on this observation, we designed a Cross-Attention module.
In this module, the text encoding serves as the Query through a fully connected layer. In contrast, the image encoding serves as the Key and Value through another fully connected layer arranged in parallel. To avoid the computational complexity of Cross-Attention, which can lead to unstable convergence, we concatenate the image encoding vector and the text encoding vector via the fully connected layer's Cutoff. Additionally, we add the output of the Cross-Attention module to construct a residual layer. This residual layer aggregates the information obtained from Cross-Attention and the feature information extracted by the original encoder, thus enhancing the overall representation.

\noindent \textbf{Integration with Residual Layers} 

\noindent The introduction of residual layers within the cross-attention framework serves multiple purposes:

\noindent \textbf{\textit{Prevention of Gradient Vanishing}} As models deepen, the risk of vanishing gradients increases, which can stall the learning process during backpropagation, especially in deep neural networks. Residual layers mitigate this by introducing skip connections that allow gradients to flow through the network more directly and efficiently.

\noindent \textbf{\textit{Enhanced Convergence}} Skip connections not only solve the vanishing gradient problem but also facilitate faster network convergence by smoothing the optimization landscape.

\noindent \textbf{\textit{Preservation of Information}} By adding a layer's input directly to its output (i.e., forming the residual), these layers ensure that no part of the network's depth loses valuable information. This aspect is crucial when integrating complex and diverse data types, as it prevents the dilution of informational content through successive processing layers.\\

\noindent \textbf{Operational Dynamics in the Model}

\noindent In practical terms, when implementing cross-attention with residuals, the output of each layer is a combination of the processed information (via cross-attention mechanisms) and the raw inputs from the preceding layer, adjusted by the network's weights. This setup enhances the model's ability to learn more effectively across layers and better generalize new, unseen data by preserving information integrity throughout the network.
This approach results in a model where each layer's output is informed by its immediate processing results and an enriched context provided by the input of previous layers. Such a configuration not only boosts the learning capacity of the model but also its ability to generalize from training data to real-world applications, making it highly effective for tasks requiring nuanced understanding and integration of multi-modal data. Assuming we have two sequence input are $X = {x_1,x_2,...,x_n}$ and $Y = {y_1,y_2,...,y_n}$. The first step of cross-attention is calculating Q(query), K(key), and V(value). The mathematical expression for calculating Q, K, and V can be described as formulas 33,34 and 35, where $W^Q$,$W^K$, and $W^V$ are learnable weight matrices.
\begin{equation}
    Q = XW^Q
\end{equation}
\begin{equation}
    K = YW^K
\end{equation}
\begin{equation}
    V = XW^V
\end{equation}
Then, the mathematical expression for calculating the attention score can be described as formula 36, where $d_k$ is the dimension of the critical vector, which is used to scale the dot product to prevent an inner product that is too large from affecting the gradient of softmax.
\begin{equation}
    A = softmax(\frac{QK^T}{\sqrt{d_k}})
\end{equation}
The mathematical expression for calculating output can be described as formula 37, where LayerNorm represents layer normalization. Adding layer normalization can help mitigate internal covariate bias, improve training stability, and accelerate model convergence.
\begin{equation}
    output = LayerNorm(AV+X)
\end{equation}
Layer normalization removes bias between batches of different input samples and maintains the consistency of input data distribution across layers by normalizing the features of each sample independently. This technique enhances model stability during training and improves the generalization and overall performance of the model.

\noindent If we only use residual links without corporating layer normalization in a fusion layer, particularly when using ReLU activation functions. It can lead to significant challenges in training deep neural networks. Here is an analysis of how these elements interact and the potential issues that can arise:

\noindent\textbf{\textit{Gradient Instability without Residual Links}} Residual links (skip connections) provide a direct pathway for gradient flow during backpropagation, which is crucial for maintaining the health of the gradient across deep networks. These connections allow gradients to bypass specific layers and merge with deeper or earlier layers, helping to mitigate issues related to gradient vanishing and explosion. Without these connections, gradients must pass through each layer sequentially, which increases the risk of vanishing (diminishing) or exploding (amplifying) gradients, especially with deep architectures.

\noindent\textbf{\textit{Impact of ReLU Activation}}The ReLU activation function, which outputs zero for negative inputs and raw input for positive inputs, can exacerbate the instability in gradient propagation. If a ReLU-activated neuron outputs a large value, and these large values stack across multiple layers without any form of moderation (like residual links or normalization), it could lead to exponentially growing activations. This scenario is often referred to as gradient explosion. The reason is that the derivatives in areas where ReLU is active are constant (i.e., 1), which can amplify gradients during backpropagation if unchecked.

\noindent\textbf{\textit{Importance of Layer Normalization}}Layer normalization can significantly stabilize the training of deep networks by normalizing the inputs across features for each data sample independently. This normalization helps control the range of activations, ensuring that the scale of outputs and gradients remains manageable across different layers. Without layer normalization, each layer can have outputs that vary widely in scale and distribution, complicating the training process and leading to unstable gradients.

\noindent\textbf{\textit{Data Disparity Across Layers}} When different layers of the network process data with significantly varying characteristics or scales (data partitioning issues), it can be challenging to maintain stable gradient flow across these layers. Residual links help by adding outputs from previous to later layers, somewhat mitigating disparities in data characteristics. However, if the data disparity is too great, even residual connections might not sufficiently stabilize the learning process, potentially leading to slow training speeds and difficulties in reaching convergence.

\noindent In summary, for architectures utilizing ReLU or similar activation functions, incorporating residual links or layer normalization (preferably both) is crucial for stabilizing training. These strategies help manage the scale of activations and gradients, ensuring more robust, efficient, and practical training of deep neural network models.
\section{Experiment and Result}
\subsection{Metrics and Running Requirements}
\subsubsection{Running Requirements}
Our hardware requirements are as follows: For running regular comparative experiments and ablation studies, you can choose CPU mode in the Colab environment and select high RAM. However, for executing the final extended comparative experiments section (Self Attention), please opt for the A100 GPU in the Colab environment and select high RAM to ensure a minimum of 64GB of memory. 
\subsubsection{Mean Accuracy}
Accuracy measures the proportion of correctly classified samples (both positive and negative) out of the total number of samples. It is calculated based on the concepts of True Positives (TP), False Positives (FP), True Negatives (TN), and False Negatives (FN). The formula for accuracy is given by formula 38:
\begin{equation}
    \text{Accuracy} = \frac{TP + TN}{TP + FP + TN + FN}
\end{equation}
Since our task involves multi-label classification, we will calculate the accuracy for each class and then compute the average to obtain the mean accuracy.
\subsubsection{F1 Score}
The F1 Score is the harmonic mean of precision and recall, providing a balanced measure of both metrics. It is particularly useful when neither precision nor recall can be prioritized over the other due to the problem domain's requirements. The formula for the F1 Score is detailed in Formula 39.
\begin{equation}
    F1 = 2 \times \frac{p \times r}{p + r}, \quad \text{where} \quad p = \frac{TP}{TP + FP}, r = \frac{TP}{TP + FN}
\end{equation}
Since our task involves multi-label classification, we will calculate the F1 for each class and then compute the average to obtain the mean F1 score.

\subsection{Comparison Experiment of Models}
The objective of the comparison experiments is to evaluate the performance of various models on a multi-label image classification task. Our strategy for model selection was to choose the best-performing model in each modality, specifically the model with the highest F1 score. The models assessed include visual models such as EfficientNet, MobileNet, and ShuffleNet, alongside natural language processing (NLP) models like ALBERT and BERT (Tiny). We utilized accuracy and F1 scores as metrics to gauge model performance across the validation datasets. The results are detailed in Table 1.
From the analysis, EfficientNet\_b4 notably excels through all the models. The data also suggest a trend where the performance of EfficientNet models tends to improve with higher version numbers. This improvement underscores the potential benefits of employing deeper or more complex network structures for handling intricate classification tasks.

\noindent Conversely, MobileNetV3 and ShuffleNet exhibit weaker performance in this task. It is likely because these models are optimized primarily for speed and operational efficiency rather than maximizing accuracy, which can lead to suboptimal results in more complex or nuanced categories. ShuffleNet demonstrates the least effectiveness among these, achieving a validation F1 score of 0.7738.

\noindent In NLP models, BERT (Tiny) generally outperforms ALBERT. This superior performance could be attributed to BERT’s enhanced capability in processing categories involving complex semantic relationships, thereby facilitating better recognition in higher-difficulty categories.

\noindent These findings provide valuable insights into the suitability of different models for specific tasks within image multi-classification, highlighting the importance of selecting a model based on the specific characteristics and demands of the classification task.

\noindent Ultimately, we selected EfficientNet\_b4 as the model for the visual modality and BERT (tiny) as the model for the natural language processing modality.

\begin{table}[H]
\centering
\begin{tabular}{|c|c|c|c|c|c|}
\hline
\makecell{Single-Modal Model}  & \makecell{Size \\ (M)} & \makecell{Time \\ Cost(s)} & \makecell{Validation \\ F1 Score} & \makecell{Validation \\ Loss} \\
\hline
Mobilenetv3\_Large\_100 & 22.1 & 44 & 0.7897 & 0.1016 \\
ShuffleNet\_v2\_x2\_0 & 28.4 & 45 & 0.7738 & 0.1035 \\
EfficientNet\_b0 & 20.5 & 51 & 0.7924 & 0.0957 \\
EfficientNet\_b1 & 31.5 & 53 & 0.7989 & 0.0931 \\
EfficientNet\_b2 & 35.2 & 50 & 0.7977 & 0.0922 \\
EfficientNet\_b3 & 49.3 & 53 & 0.8134 & 0.0875 \\
EfficientNet\_b4 & 74.5 & 57 & \textcolor{red}{0.8342} & \textcolor{red}{0.0773} \\
EfficientNetv2\_s & 96.5 & 52 & 0.8236 & 0.0834 \\
\hline
Google/Bert\_tiny & \textcolor{red}{17.6} & \textcolor{red}{40} & \textcolor{red}{0.8171} & \textcolor{red}{0.0951} \\
ALBERT & 44.1 & 48 & 0.7991 & 0.0983 \\
\hline
\end{tabular}
\caption{Single-Modal Model performance comparison}
\label{tab:model_performance}
\end{table}

\subsection{Comparsion Experiment of Parameters}
Based on the results obtained in Section 5.2, we selected the EfficientNet\_b4 model and the BERT (tiny) model for further hyperparameter analysis. Hyperparameters selected to tune for each model are learning rate, optimizer, and batch size. The evaluation metrics include validation F1 score, validation loss, training loss, and training F1 score.

\noindent We chose to tune these hyperparameters because they significantly impact the model's performance and the training process. The learning rate controls the size of the step in which the model updates the weights in each iteration. A learning rate that is too large can prevent the model from oscillating during training, making it difficult to converge; a learning rate that is too small may lead to slow convergence and long training times. A suitable learning rate can help the model converge quickly and stably, increasing training stability.

\noindent The optimizer is the algorithm that decides how to update the model's weights. A suitable optimizer can speed up convergence and improve stability, preventing the model from falling into local optima and making it easier to find the global optimum.

\noindent Batch size refers to the number of samples used in each iteration. Smaller batch sizes result in more frequent gradient updates and noisier gradient estimates, which can help the model escape local optima; larger batch sizes result in more stable gradient estimates but may be more likely to fall into local optima. The suitable batch size can balance computational efficiency and training stability, ensuring a smooth and efficient model training process.

\noindent The validation F1 score is used to evaluate the classification performance of the model on the validation set, particularly for category-imbalanced datasets. Compared to accuracy, the F1 score is a better indicator of the model's performance in such scenarios. Validation loss measures the error or loss of the model on the validation set; lower validation loss indicates better model performance on unseen data. Training loss measures the model's error or loss on the training set, reflecting how well the model fits the training data. The training F1 score assesses the classification performance of the model on the training set. It corresponds to the validation F1 score and, together with it, can determine whether the model is overfitting.
The specific validation results are shown in Tables 2 and 3 where ENet\_b4 represents model EfficientNet\_b4.
\begin{table}[H]
\centering
\setlength{\tabcolsep}{1pt}
\begin{tabular}{|c|c|c|c|c|c|c|c|c|}
\hline
Model & \makecell{Learning \\ Rate} & \makecell{Batch \\ Size} & \makecell{Optimizer} & \makecell{Time \\ Cost(s)} & \makecell{Validation \\ F1 Score} & \makecell{Validation \\ Loss} & \makecell{Train \\ Loss} & \makecell{Train \\ F1 Score} \\
\hline
ENet\_b4 & 1e-4 & 60 (max) & Adam & 57 & 0.8263 & 0.0823 & 0.0716 & 0.8906 \\
ENet\_b4 & 2e-4 & 60 (max) & Adam & 57 & 0.8355 & 0.0804 & \textcolor{red}{0.0557} & 0.9286 \\
ENet\_b4 & 1e-3 & 60 (max) & Adam & 57 & 0.8279 & 0.0826 & 0.0603 & 0.9467 \\
ENet\_b4 & 5e-4 & 60 (max) & Adam & 57 & \textcolor{red}{0.8359} & \textcolor{red}{0.0769} & 0.0601 & 0.9515 \\
ENet\_b4 & 5e-4 & 32 & Adam & \textcolor{red}{53} & 0.8321 & 0.0851 & 0.0423 & 0.9532 \\
ENet\_b4 & 5e-4 & 48 & Adam & 56 & 0.8243 & 0.0897 & 0.0967 & 0.9542 \\
ENet\_b4 & 2e-4 & 58 (max) & AdamW & 57 & 0.8353 & 0.0834 & 0.0747 & 0.9282 \\
ENet\_b4 & 5e-4 & 58 (max) & AdamW & 57 & 0.8326 & 0.0786 & 0.0415 & \textcolor{red}{0.9552} \\
ENet\_b4 & 2e-4 & 32 & AdamW & 54 & 0.8314 & 0.0794 & 0.0689 & 0.9409 \\
ENet\_b4 & 5e-4 & 32 & AdamW & 54 & 0.8349 & 0.0791 & 0.0593 & 0.9544 \\
\hline
\end{tabular}
\caption{EfficientNet\_b4 performance comparison}
\label{tab:efficientnet_b4_performance}
\end{table}

\begin{table}[H]
\centering
\setlength{\tabcolsep}{1pt}
\begin{tabular}{|c|c|c|c|c|c|c|c|c|}
\hline
Model & \makecell{Learning \\ Rate} & \makecell{Batch \\ Size} & \makecell{Optimizer} & \makecell{Time \\ Cost(s)} & \makecell{Validation \\ F1 Score} & \makecell{Validation \\ Loss} & \makecell{Train \\ Loss} & \makecell{Train \\ F1 Score} \\
\hline
BERT(Tiny) & 5e-5 & 256 & Adam & 41 & 0.8098 & 0.1078 & 0.0990 & 0.8491 \\
BERT(Tiny) & 1e-4 & 256 & Adam & \textcolor{red}{39} & 0.8103 & 0.1088 & 0.0990 & 0.8568 \\
BERT(Tiny) & 2e-4 & 256 & Adam & 40 & 0.8109 & 0.1076 & 0.1094 & 0.8373 \\
BERT(Tiny) & 5e-4 & 256 & Adam & 42 & 0.8171 & \textcolor{red}{0.0951} & 0.0943 & 0.8387 \\
BERT(Tiny) & 5e-4 & 128 & Adam & 42 & 0.8336 & 0.0981 & \textcolor{red}{0.0585} & 0.9292 \\
BERT(Tiny) & 5e-4 & 512 & Adam & 42 & \textcolor{red}{0.8339} & 0.1091 & 0.0652 & \textcolor{red}{0.9447} \\
BERT(Tiny) & 5e-4 & 256 & AdamW & 43 & 0.8183 & 0.0952 & 0.0793 & 0.8604 \\
BERT(Tiny) & 2e-4 & 256 & AdamW & 41 & 0.8141 & 0.1001 & 0.0970 & 0.8470 \\
BERT(Tiny) & 5e-4 & 128 & AdamW & 40 & 0.8064 & 0.0999 & 0.0808 & 0.8921 \\
BERT(Tiny) & 2e-4 & 128 & AdamW & 41 & 0.8161 & 0.0957 & 0.0792 & 0.8706 \\
\hline
\end{tabular}
\caption{BERT(Tiny) Performance Comparison}
\label{tab:bert_tiny_performance}
\end{table}

\noindent \textbf{Analysis of EfficientNet\_b4's Parameters: }

\noindent The comparative performance of the EfficientNet\_b4 model using different optimizers highlights critical aspects of how optimizer choice can significantly impact model training and outcomes. Table 2 shows that the Adam optimizer configuration outperforms that of the AdamW optimizer on the validation set. Specifically, the best configuration with the Adam optimizer achieves a validation F1 score of 0.8539 and a validation loss of 0.0769, compared to a validation F1 score of 0.8353 and a validation loss of 0.0834 when using the AdamW optimizer. This phenomenon may be due to that Adam is different from AdamW.

\noindent \textbf{Adam Optimizer:} Adam is known for its adaptive learning rate capabilities, which help adjust the learning rate for each parameter based on the estimates of the first and second moments of the gradients. This makes it particularly effective in scenarios where different parameters may require different learning rates or where the gradient landscape is complex and highly variable. In this case, the Adam optimizer's superior performance might be attributed to its robust handling of these adaptive adjustments, allowing for faster convergence.

\noindent \textbf{AdamW Optimizer:} While similar to Adam, AdamW modifies the update rule to incorporate weight decay directly into the optimization step rather than as a separate decay of the weights. This approach is intended to improve training in specific contexts, particularly in regularizing the model and avoiding overfitting. However, this incorporation of weight decay can sometimes lead to underutilization of the training data, especially when the decay might lead to overly penalizing the model parameters, thus potentially affecting the adaptive qualities of the learning rate adjustments.

\noindent The optimal results were achieved using a learning rate of 5e-4 and a batch size of 60. This configuration suggests that a moderately large batch size can provide more stable and reliable gradient estimates, which are crucial for practical training. Larger batches average the noise in the gradient estimations over more samples, leading to more consistent updates and potentially smoother convergence. Conversely, smaller batch sizes, while potentially faster per update and sometimes beneficial for escaping local minima, can introduce significant noise into the gradient estimates. This parameter set can result in unstable training dynamics and less efficient learning, as reflected in the lower performance metrics when smaller batches are used.
In summary, the choice of optimizer and its configuration, including learning rate and batch size, plays a critical role in the performance of deep learning models. The adaptive mechanisms of the Adam optimizer better suit the training of the EfficientNet\_b4 model under the conditions tested, leading to better performance on the validation set compared to AdamW. Moreover, selecting appropriate batch sizes further influences the stability and effectiveness of the learning process, emphasizing the need for careful experimental tuning to optimize model training and validation outcomes.

\noindent \textbf{Analysis of BERT's Parameters:}

\noindent Since we need to select models that require logit summing, our evaluation strategy prioritizes minimizing the validation loss to identify the best-performing parameter combination. Table 3 shows that the BERT (Tiny) model with the Adam optimizer configuration outperforms the AdamW optimizer on the validation set. The best-performing parameter configuration with the Adam optimizer achieves a validation F1 score of 0.8171 and a validation loss of 0.0951. In contrast, the same parameter configuration with the AdamW optimizer results in a validation F1 score of 0.8183 and a validation loss of 0.0834.

\noindent However, the AdamW optimizer's training F1 score is 0.8604, which is significantly higher than the validation F1 score, indicating overfitting. In contrast, the Adam optimizer produces consistent training and validation loss for the same parameters, demonstrating that the model performs consistently on both sets.

\noindent Lower learning rates result in slower model updates and difficulty escaping local optima, leading to poorer performance on the validation set. Higher learning rates enable faster model parameter updates, which may contribute to improved performance. Notably, a learning rate of 5e-4 with a batch size 256 achieves better model performance.

\noindent This improvement is due to larger batch sizes effectively utilizing information from the entire dataset more quickly, increasing convergence speed, and providing more stable gradient estimates. Larger batches more accurately reflect the data distribution of the entire training set, reducing the gradient's variance and leading to smoother and more stable parameter updates. It can help mitigate the instability of parameter updates at higher learning rates.

\subsection{Ablation Studies}
We designed a systematic approach for the ablation experiments to assess the impact of various model combinations and feature fusion techniques on performance. Here is an outline of the configurations and methodologies used:

\noindent \textbf{Vision Model: }This setup uses only the EfficientNet\_b4 model to process image data, providing a baseline for performance metrics using purely visual information.

\noindent \textbf{NLP Model:} Similarly, this configuration uses only BERT (Tiny) for processing textual data, offering a baseline for performance using text modalities alone.

\noindent \textbf{Fusion Model 1:} This model represents the simplest form of multimodal integration, directly combining the logits from EfficientNet\_b4 and BERT (Tiny). This method leverages the strength of both models without additional training or complexity.

\noindent \textbf{+FCNN:} This addition involves freezing the weights of both EfficientNet\_b4 and BERT (Tiny), which were pre-trained separately on their respective modalities. The feature vectors extracted from each model are then concatenated and used to train a linear output layer, effectively forming a deeper integration of the features from both modalities.

\noindent \textbf{Fusion Model 2:} Builds on the +FCNN model by incorporating logits fusion with the FCNN model outputs. This approach aims to enhance the predictive performance by integrating the linearly combined feature vectors with direct logit fusion, creating a more robust model by combining learned feature interactions and raw model predictions.

\noindent \textbf{+Cross Attention FCNN:} This configuration adds a cross-attention mechanism between the feature vectors of the two modalities. Cross-attention is an intermediate layer that focuses on the interdependencies and interactions between visual and textual features before passing them through the linear cutoff layer from the previous configurations. This method is designed to dynamically refine the focus on relevant features from each modality.

\noindent \textbf{Fusion Model 3:} Incorporates a cross-attention module equipped with residual connections and the logits from EfficientNet\_b4 and BERT (Tiny). This advanced fusion technique maximizes the synergy between the modalities, enhancing the learning dynamics and the final model performance by leveraging deep interaction layers.

\begin{table}[H]
\centering
\setlength{\tabcolsep}{1pt}
\begin{tabular}{|c|c|c|c|c|c|c|c|c|c|}
\hline
Model & \makecell{Learning \\ Rate} & \makecell{Batch \\ Size} & \makecell{Optimizer} \\
\hline
EfficientNet\_b4 & 5e-4 & 60 (max) & Adam\\
Google/Bert\_tiny & 5e-4 & 256 & Adam\\
\hline
\end{tabular}
\caption{Best Model for Vision and NLP}
\end{table}
\begin{table}[H]
\centering
\setlength{\tabcolsep}{1pt}
\begin{tabular}{|c|c|c|c|c|c|c|}
\hline
\makecell{Model \\ Modal} & \makecell{Time \\ Cost(s)} & \makecell{Validation \\ F1 Score} & \makecell{Validation \\ Loss} & \makecell{Train \\ Loss} & \makecell{Train \\ F1 Score} & \makecell{Train \\ Accuracy} \\
\hline
Vision & 61 & 0.8219 & \textcolor{red}{0.0943} & 0.0973 & 0.9345 & 0.8555 \\
NLP & \textcolor{red}{45} & 0.8358 & 0.1029 & 0.0591 & 0.8678 & \textcolor{red}{0.9387} \\
Fusion Model 1 (FM1) & 61 & - & - & - & 0.9533 & 0.8918 \\
FM1 + FCNN & 57 & 0.8421 & 0.1135 & 0.0459 & 0.9651 & 0.9156 \\
Fusion Model 2 (FM2) & 76 & - & - & - & 0.9614 & 0.9083 \\
FM2 + Cross attention FCNN & 61 & \textcolor{red}{0.8428} & 0.1530 & \textcolor{red}{0.0322} & \textcolor{red}{0.9686} & 0.9230 \\
Fusion Model 3 (FM3) & 77 & - & - & - & 0.9651 & 0.9162 \\
\hline
\end{tabular}
\caption{Ablation Studies Comparison}
\label{tab:model_performance}
\end{table}

\noindent The outcomes of these experiments are tabulated in Table 5, detailing the F1 scores, accuracy, and inference times for each model configuration. It is important to note that since Fusion Models 1,2 and 3 involve direct logits fusion without additional training, no validation F1 score or validation loss is reported.  And Logits fusion directly processes outputs from multiple models without engaging in a re-training process that typically involves loss calculations and validation set evaluations. To ensure a fair comparison of the effects of each experiment, we evaluated F1 scores, accuracy, and inference time using the entire dataset. This approach provides a comprehensive view of how each model configuration performs under consistent conditions, allowing for an accurate assessment of the efficacy of different fusion strategies and configurations in enhancing multimodal learning.
\subsection{Final Model with Performance}
For our multimodal learning system, we strategically selected and configured different models for the visual and textual modalities to achieve exceptional performance as evidenced by our Kaggle validation scores.
\subsubsection{Visual Modality Configuration:}
\textbf{Model:} EfficientNet\_b4 was selected due to its robust performance in image classification tasks.
Learning Rate: Set at 5e-4, this rate provides an optimal balance between speed and accuracy in the model's convergence.

\noindent \textbf{Batch Size:} We used a batch size of 60, which offers a good compromise between computational efficiency and the stability of gradient estimations.

\noindent \textbf{Optimizer: }The Adam optimizer was employed for its adaptive learning rate capabilities, which enhance the ability to converge faster and more reliably.
\subsubsection{Textual Modality Configuration:}
\textbf{Model:} BERT (Tiny) was chosen for its efficiency and effectiveness in handling natural language processing tasks, making it particularly suitable for environments where computational resources and model latency are considerations.
\noindent \textbf{Learning Rate:} It is also set to 5e-4, mirroring the setup of the visual modality to maintain consistency in training dynamics across modalities.

\noindent \textbf{Batch Size:} For BERT (Tiny), a larger batch size 256 was selected to facilitate smoother gradient calculations over more extensive data samples.

\noindent \textbf{Optimizer: }Similar to the visual modality, the Adam optimizer was utilized to take advantage of its adaptive learning rate mechanisms.
\subsubsection{Fusion Module Configuration:}
\textbf{Approach:} We implemented a cross-attention mechanism with an additional residual module in the fusion layer. This setup allows the model to focus more effectively on relevant features from both modalities, enhancing the integration and interaction of visual and textual data.

\noindent \textbf{Training Strategy:} As detailed in Section 4.1.4, a semi-supervised training approach was adopted. This method leverages unlabeled data alongside labeled examples, effectively enlarging the training dataset and improving the model's generalization capabilities.

\noindent \textbf{Performance and Outcome:}
 Our model scored 0.96641 validation Score on Kaggle. And it is the highest-performing submission among all our entries. This score reflects the efficacy of the chosen models and their configurations and underscores the success of our fusion strategy and training approach. This comprehensive setup demonstrates the power of combining advanced deep learning techniques with strategic model configuration and innovative data fusion approaches. The high validation score is a testament to the effectiveness of integrating multiple modalities through sophisticated neural architectures and training strategies.
\section{Conclusion}
This study presents a comprehensive approach to multi-label image classification by integrating advanced visual and textual feature learning within a multi-modal framework. By leveraging the strengths of visual models based on Convolutional Neural Networks for image processing and Natural Language Processing (NLP) models for textual analysis, the proposed Model effectively enhances the accuracy and efficiency of label prediction. Combining visual and textual features and incorporating a fusion module further improves the Model's capability to handle complex multi-label classification tasks.

\noindent The experimental results demonstrate that EfficientNet\_b4 and BERT (Tiny) models, selected for their superior performance, significantly outperform other state-of-the-art models in visual and textual modalities. Adopting data augmentation techniques, weighted loss functions, and a robust label assignment strategy ensures the Model's resilience and accuracy, particularly in imbalanced datasets. Additionally, the innovative use of semi-supervised learning and cross-attention mechanisms with residual layers effectively enhances the Model's generalization and stability.

\noindent This work contributes to the field by providing a scalable and efficient solution for multi-label image classification suitable for deployment in resource-constrained environments. Future research can explore integrating additional modalities and refining fusion techniques to advance multi-modal classifiers' capabilities further.
\subsection{Future Work}
For future enhancements of our multimodal learning model, exploring more sophisticated fusion techniques and dynamically adaptive mechanisms can significantly refine how modalities interact and contribute to the overall predictive power of the system. Below specifies potential advancements.

\subsubsection{Dynamic Weighting of Modalities:}
\textbf{Propose: }Instead of statically combining features from different modalities, dynamic weighting adjusts the influence of each modality based on its relevance to the prediction task at hand.

\noindent \textbf{Implementation:} Implementing a learnable weighting system that evaluates the context or the specific input instance to determine the contribution weight of each modality. For instance, in scenarios where visual data is more descriptive than textual data, the model could dynamically increase the visual modality's weight.
\subsubsection{Adaptive Attention Mechanisms:}
\textbf{Propose: }To enhance the model's ability to focus selectively on the most informative parts of the data from each modality.

\noindent \textbf{Implementation: } Techniques such as transformer models or variants can be utilized, where multi-head attention mechanisms can be tailored to multimodal data. Each 'head' in the attention mechanism could learn to attend to different aspects of the modalities, dynamically focusing on more relevant features.

\subsubsection{Contextual Modality Adjustment:} 
\textbf{Purpose: } To modify the contribution of each modality based on the situational context or inherent data characteristics.

\noindent \textbf{Implementation: } Introducing a context analysis layer that assesses external factors or embedded cues within the data that may signal the appropriateness of relying more heavily on one modality over another. For example, the model might lean more on textual descriptions in noisy image conditions.
\subsubsection{Enhanced Fusion Techniques:}
\textbf{Purpose: } To develop more complex fusion strategies beyond simple concatenation or addition of features.

\noindent \textbf{Implementation: } Exploring architectures like gated multimodal units where gates control the flow and integration of modal information based on the input state or using capsule networks that can maintain hierarchical relationships between features from different modalities.
\subsubsection{Incremental Learning and Feedback Mechanisms:}
\textbf{Purpose: } To allow the model to refine its understanding and handling of multimodal data over time, particularly in changing environments or as more data becomes available.

\noindent \textbf{Implementation: } Incorporating feedback loops where the model's predictions and outcomes inform subsequent adjustments to modal weights or attention focuses, akin to reinforcement learning strategies.

\noindent By integrating these advanced techniques, our model can not only improve in terms of raw performance metrics like accuracy or F1 score but also robustness, adaptability, and interpretability, making it more effective in practical, real-world applications where the dynamics of modalities can be quite complex.


\bibliographystyle{ieeetrans}
\bibliography{Assignment_Ref}

\section*{Appendix}
\subsection*{Running code}

You can access our code at the following link: \href{https://colab.research.google.com/drive/1IeTkVXBFNo8d78Gj0sx9PuFs0dTfOvcU?usp=sharing}{colab link}\\
You can access our data at the following link: 
\href{https://drive.google.com/file/d/1kHIw0thiBJHi5DNb_VRi0xwcoxLUtMOG/view?usp=drive_link}{Predicted labels 1}, 
\href{https://drive.google.com/file/d/1Q2pVnn0zLCiD3NtF1Te0YXEvytMe4i7b/view?usp=drive_link}{Predicted labels 2}\\
Then you must change the \textbf{runtime type to A100}
\begin{figure}[H]
        \centering
        \includegraphics[width=\textwidth]{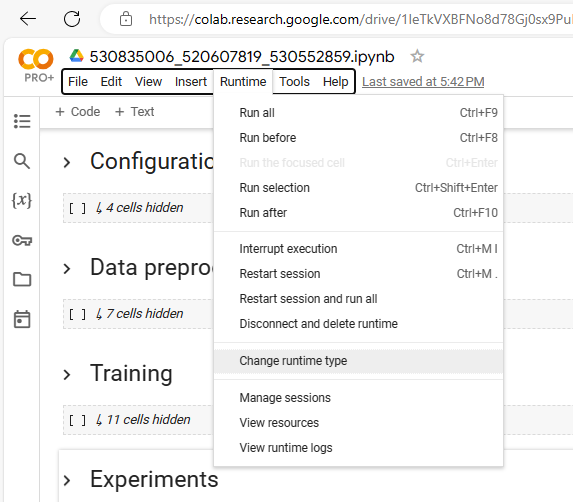}
        \caption{Change runtime type}
\end{figure}
\begin{figure}[H]
        \centering
        \includegraphics[width=\textwidth]{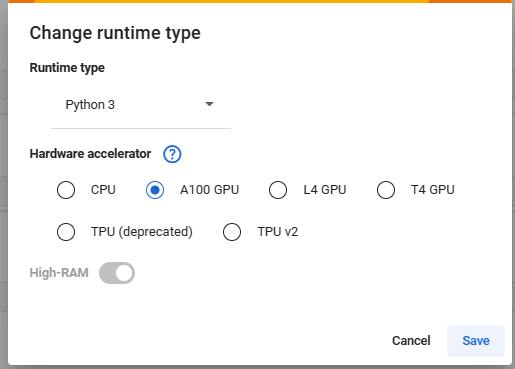}
        \caption{Choose A100}
\end{figure}
The structure of the code is shown below, first you need to run \textbf{Configuration environment, Data preprocessing, Training}. 

\begin{figure}[H]
        \centering
        \includegraphics[width=\textwidth]{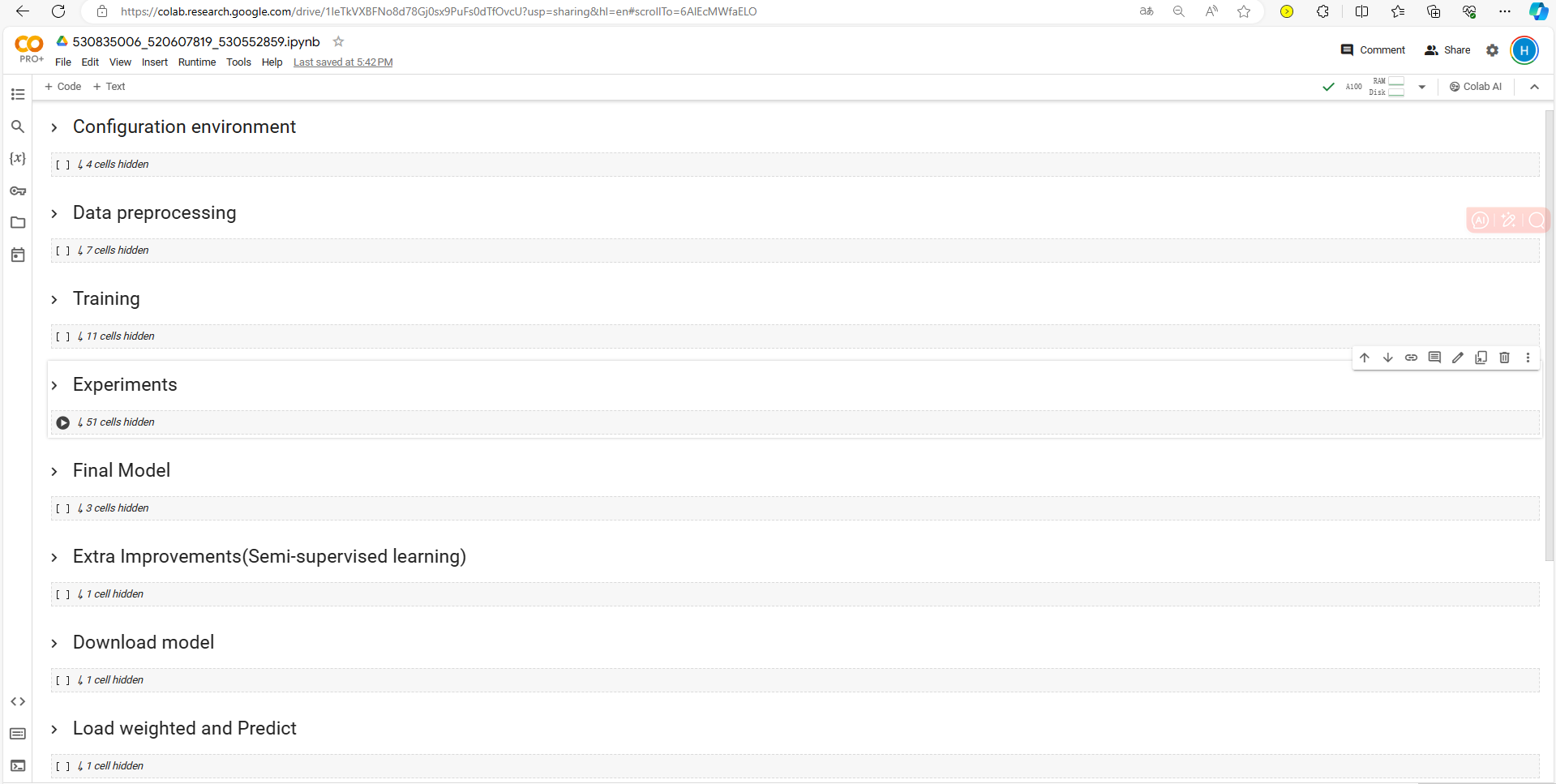}
        \caption{Code structure}
\end{figure}
\newpage
\textbf{Then}

1)if you want to \textbf{check the results, run Load weighted and Predict} directly. You can download \textbf{Submission.csv} directly to evaluate our score!

2)If you want to see the \textbf{training process of the final model, run Final Model} directly, and the weights file and CSV will be saved directly after the model is trained.(The model obtained scores around 0.955 on the leader-boards)

3)If you want to check the results of semi-supervised training, run Final Model and Extra Improvements(The model obtained scores around 0.966 on the leader-boards)

\textbf{Warning:}

1)The data for the experiment has been saved in the code block, if you need to check the data for the experiment, please do not run Experiments as this will clear the saved data. If you want to check if the experiment is runnable, run this part of the code, but it will take a long time.

2)Running Download model will download the optimal model obtained from ablation experiments and semi-supervised learning, but if you have not run Experiments, Final Model, Extra Improvements, please do not run this module. It will report an error because of the missing model!

\newpage
\subsection*{Appendix-Model Comparison}
\begin{table}[H]
\centering
\begin{tabular}{|c|c|c|c|c|c|}
\hline
Model & F1 Score & Accuracy & Class1\_F1 & Class2\_F1 & Class3\_F1\\
\hline
EfficientNet\_B0 & 0.899279 & 0.7749 & 0.9692 & 0.7508 & 0.8329 \\
EfficientNet\_B1 & 0.913134 & 0.8028 & 0.9754 & 0.7054 & 0.8490\\
EfficientNet\_B2 & 0.914415 & 0.8063 & 0.9739 & 0.7957 & 0.8579\\
EfficientNet\_B3 & 0.923831 & 0.822867 & 0.9742 & 0.8320 & 0.8252\\
EfficientNet\_B4 & 0.952135 & 0.8798 & 0.9870 & 0.8683 & 0.9268\\
MobileNetV3\_large\_100 & 0.883465 & 0.7478 & 0.9643 & 0.7042 & 0.7432\\
ShuffleNet & 0.867072 & 0.723367 & 0.9566 & 0.6794 & 0.7917\\
ALBERT-base-V2 & 0.818616 & 0.674733 & 0.9386 & 0.4767 & 0.4711\\
BERT(Tiny) & 0.838747 & 0.6945 & 0.9481 & 0.5385 & 0.5215\\
\hline
\end{tabular}
\caption{Model Comparison on Train Loss(1/4)}
\end{table}

\begin{table}[H]
\centering
\begin{tabular}{|c|c|c|c|c|c|}
\hline
Model & Class4\_F1 & Class5\_F1 & Class6\_F1 & Class7\_F1 & Class8\_F1\\
\hline
EfficientNet\_B0 & 0.8781 & 0.9596 & 0.8692 & 0.9530 & 0.7299 \\
EfficientNet\_B1  & 0.9095 & 0.9786 & 0.7838 & 0.9623 & 0.7713 \\
EfficientNet\_B2  & 0.8651 & 0.9558 & 0.8765 & 0.9508 & 0.7467\\
EfficientNet\_B3  & 0.9366 & 0.9810 & 0.9098 & 0.9643 & 0.8278\\
EfficientNet\_B4  & 0.9462 & 0.9880 & 0.9228 & 0.9815 & 0.8433\\
MobileNetV3\_large\_100  & 0.8257 & 0.9661 & 0.8296 & 0.9466 & 0.7070 \\
ShuffleNet  & 0.7959 & 0.9502 & 0.8086 & 0.8916 & 0.6852\\
ALBERT-base-V2  & 0.7833 & 0.9689 & 0.7747 & 0.9205 & 0.4545 \\
BERT(Tiny)  & 0.8126 & 0.9727 & 0.7805 & 0.9264 & 0.5024 \\
\hline
\end{tabular}
\caption{Model Comparison on Train Loss(2/4)}
\end{table}

\begin{table}[H]
\centering
\begin{tabular}{|c|c|c|c|c|c|}
\hline
Model  & Class9\_F1 & Class10\_F1 & Class11\_F1 & Class12\_F1 & Class13\_F1 \\
\hline
EfficientNet\_B0 & 0.8679 & 0.8005 & 0.8111 & 0.7879 & 0.7092 \\
EfficientNet\_B1 & 0.9035 & 0.8428 & 0.8571 & 0.8364 & 0.8398 \\
EfficientNet\_B2  & 0.8846 & 0.8389 & 0.8218 & 0.8197 & 0.8063 \\
EfficientNet\_B3  & 0.9260 & 0.8675 & 0.8497 & 0.8362 & 0.8017 \\
EfficientNet\_B4  & 0.9456 & 0.9201 & 0.8617 & 0.8590 & 0.8584 \\
MobileNetV3\_large\_100 & 0.8865 & 0.7982 & 0.7647 & 0.7921 & 0.6488 \\
ShuffleNet  & 0.8222 & 0.7029 & 0.7859 & 0.7890 & 0.5543 \\
ALBERT-base-V2 & 0.7391 & 0.5194 & 0.7513 & 0.6615 & 0.6788 \\
BERT(Tiny) & 0.7762 & 0.5321 & 0.7505 & 0.6617 & 0.6943 \\
\hline
\end{tabular}
\caption{Model Comparison on Train Loss(3/4)}
\end{table}

\begin{table}[H]
\centering
\begin{tabular}{|c|c|c|c|c|c|}
\hline
Model & Class14\_F1 & Class15\_F1 & Class16\_F1 & Class17\_F1 & Class18\_F1 \\
\hline
EfficientNet\_B0 & 0.6749 & 0.7915 & 0.8977 & 0.8141 & 0.9257 \\
EfficientNet\_B1 & 0.6932 & 0.8460 & 0.9416 & 0.8586 & 0.9033 \\
EfficientNet\_B2 & 0.7513 & 0.8251 & 0.9400 & 0.8635 & 0.9345 \\
EfficientNet\_B3 & 0.8079 & 0.8235 & 0.9488 & 0.8432 & 0.9475 \\
EfficientNet\_B4 & 0.8348 & 0.8954 & 0.9755 & 0.9393 & 0.9491 \\
MobileNetV3\_large\_100 & 0.5886 & 0.7634 & 0.8881 & 0.8121 & 0.8766 \\
ShuffleNet & 0.5324 & 0.7524 & 0.8844 & 0.7457 & 0.8566 \\
ALBERT-base-V2 & 0.4378 & 0.7841 & 0.9456 & 0.8614 & 0.9359 \\
BERT(Tiny) & 0.5105 & 0.7988 & 0.9438 & 0.8699 & 0.9438 \\
\hline
\end{tabular}
\caption{Model Comparison on Train Loss(4/4)}
\end{table}

\subsection*{Appendix-Parameter Comparison}
\begin{table}[H]
\centering
\setlength{\tabcolsep}{1pt}
\begin{tabular}{|c|c|c|c|c|c|c|c|}
\hline
Model & Param & F1 Score & Accuracy & Class1\_F1 & Class2\_F1 & Class3\_F1\\
\hline
EfficientNetB4 & Learning Rate:1e-4 & 0.8906 & 0.7616 & 0.9715 & 0.6585 & 0.7917 \\
EfficientNetB4 & Learning Rate:2e-4 & 0.9286 & 0.8300 & 0.9840 & 0.7761 & 0.8657 \\
EfficientNetB4 & Learning Rate:1e-3 & 0.9467 & 0.8674 & 0.9836 & 0.8689 & 0.9130 \\
EfficientNetB4 & Batch Size:32 & 0.9532 & 0.8820 & 0.9859 & 0.8546 & 0.9283 \\
EfficientNetB4 & Batch Size:48 & 0.9543 & 0.8900 & 0.9892 & 0.8687 & 0.8742 \\
EfficientNetB4 & Optimizer:AdamW & 0.9552 & 0.8856 & 0.9880 & 0.8708 & 0.9271 \\
BERT(Tiny) & Learning Rate:2e-4 & 0.8373 & 0.6982 & 0.9504 & 0.4728 & 0.5406 \\
BERT(Tiny) & Learning Rate:1e-4 & 0.8568 & 0.7209 & 0.9611 & 0.5694 & 0.6085 \\
BERT(Tiny) & Learning Rate:5e-5 & 0.8491 & 0.7135 & 0.9585 & 0.4819 & 0.5708 \\
BERT(Tiny) & Batch Size:128 & 0.9292 & 0.8500 & 0.9766 & 0.7637 & 0.7851 \\
BERT(Tiny) & Batch Size:512 & 0.9447 & 0.8781 & 0.9818 & 0.8382 & 0.8395 \\
BERT(Tiny) & Optimizer:AdamW & 0.8604 & 0.7259 & 0.9574 & 0.6062 & 0.5944 \\
\hline
\end{tabular}
\caption{Parameter comparison on Train Loss(1/4)}
\end{table}

\begin{table}[H]
\centering
\setlength{\tabcolsep}{1pt}
\begin{tabular}{|c|c|c|c|c|c|c|}
\hline
Model & Param Name  & Class4\_F1 & Class5\_F1 & Class6\_F1 & Class7\_F1 & Class8\_F1 \\
\hline
EfficientNetB4 & Learning Rate:1e-4 & 0.8615 & 0.9735 & 0.8469 & 0.9476 & 0.6820 \\
EfficientNetB4 & Learning Rate:2e-4 & 0.9103 & 0.9862 & 0.9030 & 0.9663 & 0.7861 \\
EfficientNetB4 & Learning Rate:1e-3 & 0.9518 & 0.9841 & 0.9187 & 0.9737 & 0.8730 \\
EfficientNetB4 & Batch Size:32 & 0.9364 & 0.9884 & 0.9286 & 0.9744 & 0.8717 \\
EfficientNetB4 & Batch Size:48 & 0.9500 & 0.9853 & 0.9365 & 0.9757 & 0.8762 \\
EfficientNetB4 & Optimizer:AdamW & 0.9487 & 0.9898 & 0.9284 & 0.9797 & 0.8828 \\
BERT(Tiny) & Learning Rate:2e-4 & 0.8076 & 0.9702 & 0.7808 & 0.9261 & 0.4738 \\
BERT(Tiny) & Learning Rate:1e-4 & 0.8254 & 0.9771 & 0.7847 & 0.9269 & 0.5816 \\
BERT(Tiny) & Learning Rate:5e-5 & 0.8283 & 0.9745 & 0.7787 & 0.9285 & 0.5254 \\
BERT(Tiny) & Batch Size:128 & 0.9251 & 0.9922 & 0.9065 & 0.9734 & 0.7938 \\
BERT(Tiny) & Batch Size:512 & 0.9467 & 0.9942 & 0.9213 & 0.9786 & 0.8320 \\
BERT(Tiny) & Optimizer:AdamW & 0.8476 & 0.9794 & 0.8091 & 0.9332 & 0.6257 \\
\hline
\end{tabular}
\caption{Parameter comparison on Train Loss(2/4)}
\end{table}

\begin{table}[H]
\centering
\setlength{\tabcolsep}{1pt}
\begin{tabular}{|c|c|c|c|c|c|c|}
\hline
Model & Param Name & Class9\_F1 & Class10\_F1 & Class11\_F1 & Class12\_F1 & Class13\_F1 \\
\hline
EfficientNetB4 & Learning Rate:1e-4 & 0.8775 & 0.7477 & 0.7527 & 0.7343 & 0.6368 \\
EfficientNetB4 & Learning Rate:2e-4 & 0.9385 & 0.8122 & 0.8183 & 0.8162 & 0.8121 \\
EfficientNetB4 & Learning Rate:1e-3 & 0.9152 & 0.9051 & 0.8778 & 0.8523 & 0.8349 \\
EfficientNetB4 & Batch Size:32 & 0.9622 & 0.8954 & 0.8994 & 0.8531 & 0.8787 \\
EfficientNetB4 & Batch Size:48 & 0.9584 & 0.9192 & 0.8818 & 0.8772 & 0.8817 \\
EfficientNetB4 & Optimizer:AdamW & 0.9544 & 0.9132 & 0.8730 & 0.8603 & 0.8874 \\
BERT(Tiny) & Learning Rate:2e-4 & 0.7582 & 0.4485 & 0.7250 & 0.6624 & 0.4773 \\
BERT(Tiny) & Learning Rate:1e-4 & 0.7811 & 0.5599 & 0.7435 & 0.6180 & 0.6488 \\
BERT(Tiny) & Learning Rate:5e-5 & 0.7719 & 0.5486 & 0.7409 & 0.5694 & 0.6684 \\
BERT(Tiny) & Batch Size:128 & 0.9018 & 0.7920 & 0.9073 & 0.8859 & 0.8812 \\
BERT(Tiny) & Batch Size:512 & 0.9318 & 0.8301 & 0.9127 & 0.9034 & 0.9015 \\
BERT(Tiny) & Optimizer:AdamW & 0.8107 & 0.5315 & 0.7852 & 0.7068 & 0.7368 \\
\hline
\end{tabular}
\caption{Parameter comparison on Train Loss(3/4)}
\end{table}

\begin{table}[H]
\centering
\setlength{\tabcolsep}{1pt}
\begin{tabular}{|c|c|c|c|c|c|c|}
\hline
Model & Param Name & Class14\_F1 & Class15\_F1 & Class16\_F11 & Class17\_F1 & Class18\_F1 \\
\hline
EfficientNetB4 & Learning Rate:1e-4 & 0.5321 & 0.8050 & 0.9376 & 0.8492 & 0.9100 \\
EfficientNetB4 & Learning Rate:2e-4 & 0.7069 & 0.8645 & 0.9621 & 0.9045 & 0.9540 \\
EfficientNetB4 & Learning Rate:1e-3 & 0.8009 & 0.8924 & 0.9647 & 0.9161 & 0.9640 \\
EfficientNetB4 & Batch Size:32 & 0.8671 & 0.9016 & 0.9737 & 0.9297 & 0.9621 \\
EfficientNetB4 & Batch Size:48 & 0.8248 & 0.9221 & 0.9714 & 0.9307 & 0.9624 \\
EfficientNetB4 & Optimizer:AdamW & 0.8458 & 0.9062 & 0.9757 & 0.9361 & 0.9730 \\
BERT(Tiny) & Learning Rate:2e-4 & 0.5154 & 0.8037 & 0.9408 & 0.8740 & 0.9353 \\
BERT(Tiny) & Learning Rate:1e-4 & 0.5599 & 0.8158 & 0.9528 & 0.8898 & 0.9528 \\
BERT(Tiny) & Learning Rate:5e-5 & 0.5218 & 0.8149 & 0.9465 & 0.8842 & 0.9435 \\
BERT(Tiny) & Batch Size:128 & 0.7740 & 0.9185 & 0.9774 & 0.9469 & 0.9825 \\
BERT(Tiny) & Batch Size:512 & 0.8225 & 0.9359 & 0.9845 & 0.9646 & 0.9860 \\
BERT(Tiny) & Optimizer:AdamW & 0.5594 & 0.8370 & 0.9551 & 0.8973 & 0.9483 \\
\hline
\end{tabular}
\caption{Parameter comparison on Train Loss(4/4)}
\end{table}

\subsection*{Appendix-Ablation Studies}
\begin{table}[H]
\centering
\setlength{\tabcolsep}{1pt}
\begin{tabular}{|c|c|c|c|c|c|c|c|}
\hline
Model & F1 Score & Accuracy & Class1\_F1 & Class2\_F1 & Class3\_F1 \\
\hline
Vision & 0.9346 & 0.8555 & 0.9778 & 0.8020 & 0.8703 \\
NLP & 0.9387 & 0.8678 & 0.9805 & 0.8199 & 0.8216 \\
Logits Fusion(Vision\&NLP) & 0.9533 & 0.8918 & 0.9879 & 0.8273 & 0.9015 \\
+FCNN & 0.9651 & 0.9156 & 0.9908 & 0.8843 & 0.9353 \\
Logits Fusion(+FCNN) & 0.9651 & 0.9162 & 0.9903 & 0.8884 & 0.9241 \\
+Cross attention FCNN & 0.9686 & 0.9230 & 0.9911 & 0.8980 & 0.9378 \\
Logits Fusion(+Cross attention FCNN) & 0.9651 & 0.9162 & 0.9903 & 0.8884 & 0.9276 \\
\hline
\end{tabular}
\caption{Ablation Studieson Train Loss(1/4)}
\end{table}

\begin{table}[H]
\centering
\setlength{\tabcolsep}{1pt}
\begin{tabular}{|c|c|c|c|c|c|c|}
\hline
Model & Class4\_F1 & Class5\_F1 & Class6\_F1 & Class7\_F1 & Class8\_F1 \\
\hline
Vision & 0.9242 & 0.9716 & 0.9073 & 0.9672 & 0.7913 \\
NLP & 0.9420 & 0.9848 & 0.9082 & 0.9748 & 0.7910 \\
Logits Fusion(Vision\&NLP) & 0.9521 & 0.9894 & 0.9361 & 0.9795 & 0.8068 \\
+FCNN & 0.9549 & 0.9928 & 0.9426 & 0.9831 & 0.8871 \\
Logits Fusion(+FCNN) & 0.9570 & 0.9918 & 0.9392 & 0.9805 & 0.8626 \\
+Cross attention FCNN & 0.9574 & 0.9921 & 0.9523 & 0.9857 & 0.8938 \\
Logits Fusion(+Cross attention FCNN) & 0.9570 & 0.9921 & 0.9470 & 0.9847 & 0.8650 \\
\hline
\end{tabular}
\caption{Ablation Studies on Train Loss(2/4)}
\end{table}

\begin{table}[H]
\centering
\setlength{\tabcolsep}{1pt}
\begin{tabular}{|c|c|c|c|c|c|c|}
\hline
Model & Class9\_F1 & Class10\_F1 & Class11\_F1 & Class12\_F1 & Class13\_F1 \\
\hline
Vision & 0.9324 & 0.8595 & 0.8655 & 0.8458 & 0.8577 \\
NLP & 0.9298 & 0.8194 & 0.8974 & 0.8768 & 0.8672 \\
Logits Fusion(Vision\&NLP) & 0.9549 & 0.8924 & 0.8932 & 0.8606 & 0.8812 \\
+FCNN & 0.9663 & 0.8997 & 0.8967 & 0.8848 & 0.8757 \\
Logits Fusion(+FCNN) & 0.9664 & 0.9114 & 0.9205 & 0.8899 & 0.8800 \\
+Cross attention FCNN & 0.9681 & 0.9133 & 0.9323 & 0.8958 & 0.9055 \\
Logits Fusion(+Cross attention FCNN) & 0.9664 & 0.9114 & 0.9205 & 0.8899 & 0.8993 \\
\hline
\end{tabular}
\caption{Ablation Studies on Train Loss(3/4)}
\end{table}

\begin{table}[H]
\centering
\setlength{\tabcolsep}{1pt}
\begin{tabular}{|c|c|c|c|c|c|}
\hline
Model & Class14\_F1 & Class15\_F1 & Class16\_F1 & Class17\_F1 & Class18\_F1 \\
\hline
Vision & 0.7639 & 0.8766 & 0.9600 & 0.9196 & 0.9466 \\
NLP & 0.8141 & 0.9319 & 0.9816 & 0.9591 & 0.9772 \\
Logits Fusion(Vision\&NLP) & 0.8144 & 0.9106 & 0.9843 & 0.9586 & 0.9773 \\
+FCNN & 0.8796 & 0.9263 & 0.9880 & 0.9698 & 0.9817 \\
Logits Fusion(+FCNN) & 0.8565 & 0.9213 & 0.9868 & 0.9677 & 0.9812 \\
+Cross attention FCNN & 0.8948 & 0.9450 & 0.9897 & 0.9779 & 0.9875 \\
Logits Fusion(+Cross attention FCNN) & 0.8751 & 0.9370 & 0.9924 & 0.9756 & 0.9871 \\
\hline
\end{tabular}
\caption{Ablation Studies on Train Loss(4/4)}
\end{table}

\end{document}